# The computerization of archaeology: survey on AI techniques

**Lorenzo Mantovan[1] *, Loris Nanni[2]**

[1] DEI - University of Padova, Via Gradenigo, 6 - 35131- Padova, Italy.
Email: lorenzo.mantovan@studenti.unipd.it
[2] DEI - University of Padova, Via Gradenigo, 6 - 35131- Padova, Italy.
EMAIL: loris.nanni@unipd.it
*Corresponding Author: Lorenzo Mantovan, Email: lorenzo.mantovan@gmail.com

## Abstract

This paper analyses the application of artificial intelligence techniques to various areas of archaeology and more specifically:

- The use of software tools as a creative stimulus for the organization of exhibitions; the use of humanoid robots and holographic displays as guides that interact and involve museum visitors;
- The analysis of methods for the classification of fragments found in archaeological excavations and for the reconstruction of ceramics, with the recomposition of the parts of text missing from historical documents and epigraphs;
- The cataloguing and study of human remains to understand the social and historical context of belonging with the demonstration of the effectiveness of the AI techniques used;
- The detection of particularly difficult terrestrial archaeological sites with the analysis of the architectures of the Artificial Neural Networks most suitable for solving the problems presented by the site; the design of a study for the exploration of marine archaeological sites, located at depths that cannot be reached by man, through the construction of a freely explorable 3D version.

## 1. Introduction

In classical Greek, *archaiologhia* is a word composed of two elements: *archaios*, which means ancient, and *loghia*, that is speech; in the early nineteenth century the term archaeology assumed the meaning of studying the finds of antiquity in their artistic meaning: the sculptures, paintings and ancient buildings seen in relation to the categories of beauty and ugliness, without taking into account the context in which the artefact it had been produced. In the same period, the first major European public museums were established, such as the Louvre in Paris, which generally contained an archaeological section with bas-reliefs and ancient statues.

Over time, the word archaeology, however, acquired a more autonomous meaning from the history of classical art: it went on to indicate the excavation activity, carried out through specific techniques, for the search for finds from past civilizations. Nowadays archaeology deals, with the same interest and with the same passion, both of a "miraculously" intact statue and of a simple crock of vase. Over the years, a new and more correct idea has matured according to which the finding is studied to have more information on the historical-political, economic and cultural condition of the period to which it belongs. Archaeology also makes use of other disciplines defined, for this, "related" or "auxiliary" such as archeometry, anthropology, remote sensing and numismatics [1].

With the development of technology and the spread of information technology in many scientific and humanistic disciplines, including archaeology, computer archaeology is born. It is an area of archaeological research that promotes the formalized representation of knowledge, making use of IT methods and techniques to acquire, process and transmit data. Computer archaeology, now often referred to as digital archaeology, has its own field of investigation which involves the sectors traditionally aimed at field research, data processing in the laboratory, management, protection and enhancement of archaeological heritage [2]. Systematic publications, such as the *Proceedings of the annual Computer applications and quantitative methods in archaeology* [3] and the magazine *«Archeologia e calcolatori»* [4], allow an updated assessment of scientific activity.

The main areas in which archaeological and computer science knowledge intertwine to provide simple and fast methods for scholars to understand the nature of the finds, to facilitate research and study operations in areas not accessible by man and to make more interactive visits to the museum.

Each sections analyses a problem that afflicts archaeological analysis and lists some projects that aim to solve it: section 2 deals with the topic on how to renew the museum environment and how to interact and visit an exhibition; section 3 focuses on how fragmentary artefacts and texts can be reconstructed; section 4 proposes computer techniques for the analysis of human skeletons found and, finally, section 5 exposes software tools useful for the study and meeting of archaeological sites. In the **Table 1**, all the methods that will be exposed are summarized and compared.

| Archeological field | Name | Summary | Artificial intelligence technique used |
|---|---|---|---|
| **Museum area** | Robovie-R ver.2 | It is a sociological project that aims at a humanoid robot that simulates the explanation of a work with movements similar to those of a real guide. | Face recognition, future implementation of response methods with AI. |
| | RHINO | It is a robot who plans interactive tours of the museum, he is also capable of calculating routes in crowded spaces. | Localization of Markov, Backpropagation network for the creation of occupancy maps, μDWA, value iteration. |
| | Minerva | It is a software that allows you to group artefacts according to some criteria provided by the user and arrange them in the various rooms of the museum respecting the safety rules. | Multi-agent system developed in DAI; the agents are: preparator, allocator, navigator, commentator. |
| | Interactive and holographic system for exhibitions | It is an exhibitor that uses holographic technique and other sensors to display and interact with the models represented; equipped with software capable of interpreting users' questions and processing answers. | Watson Natural Language (NLC), Face recognition. |
| **Reconstruction of finds** | ArchAIDE | It is a tool that allows the recognition and storage in the database of the ceramic fragments found, there is also a version for smartphones and tablets. | Recognition system obtained by training a Convolutional Neural Network (CNN). |
| | Pairing and reconstruction of fragments | It is a software that allows you to group similar fragments of ostraka and subsequently for each group it proposes a series of possible reconstructions. | Siamese Neural Network (particular type of CNN). |
| | Texts and epigraph | It is a software that receives input parts of inscriptions of epigraphs and returns the 20 possible reconstructions of the damaged text. | PYTHIA, Long-Short Term Memory (LSTM). |
| **Study of human remains** | Determination sex from skull | It is software that receives an image of a skull as input and, after various processing, returns membership in the "Male" or "Female" class. | GoogLeNet (CNN trained in image recognition). |
| | Calculating stature from bones | Two software are compared that receive the measurements of the bones of the skeleton as input and return an estimate of the individual. | Genetic Algorithm (GA), Artificial Neural Network (ANN). |
| **Identification of archeological sites** | Terrestrial sites | After a brief introduction on Archaeological predictive modelling (APM), four projects are exposed: The first is the search for archaeological sites in Brandenburg using Ainet, a software based on PNN and SOM; The second is the search for archaeological sites of tombs in the Eurasian steppes; the third is a study carried out with GPR on the city of Falerii Novi and the fourth is the study of the monumental site of Kuelap using LiDAR and AI. | Multilayer Perceptron Network (MLP), Probabilistic Neural Network (PNN), Convolutional Neural Network (CNN), Self-Organizing Feature Map (SOM). |
| | VENUS | It is a project that aims to create marine archaeological sites with augmented reality, starting from the data collected by autonomous marine or remote-control vehicles. | Software tools for 3D data processing and consistency check between them. |

**Table 1:** *Summary table of all the methods covered in the overview.*

## 2. Algorithms of artificial intelligence in the museum area

Many cultural events and museums, in recent years, have tried to integrate new technologies (cutting-edge visualization and interaction technologies) to develop the spread of culture and make visitors of all ages more curious and interested. Artificial intelligence, in this environment, is mainly used by:

- Chatbots and robots both to understand the questions asked by people and to elaborate adequate answers, and to move and elaborate paths in the museum [5], [6], [8].

- Support software for the design of exhibitions [19]–[21], [23].

### 2.1. Interactive guide robots

Numerous "intelligent" and interactive robot projects allow visitors to participate in guided tours and feel involved during the exhibition of the works, but the projects and studies conducted on Robovie-R ver.2 [5] and RHINO [6] are noteworthy.

#### 2.1.1. Robovie-R ver.2

In Japan, a team from the University of Saitama, led by Yoshinori Kuno, carried out a sociological study examining the "friendly" human-robot interactions through non-verbal gestures [5], [8]. The human behaviors of the guides were observed during the explanation to the visitors and, in particular, the moments of the explanation in which the guide turned his head towards the visitors were examined. From the **Table 2**, you can see the number of occurrences and when it occurred.

| Moments of explanation | Number of occurrences |
|---|---|
| TRP (transition relevance place) | 61 |
| When saying keywords with emphasis | 14 |
| When saying unfamiliar words or citing figures | 6 |
| When using deictic such as this | 23 |
| With hand gestures | 41 |
| When the visitors asked questions | 12 |

**Table 2:** *It collects the number of cases in which the guide turns his head and the number of times that occurred during the exhibition of a piece of work.*

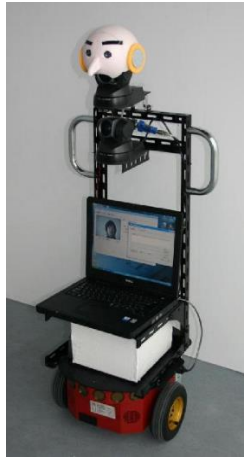

**Figure 1:** *The prototype of the project is a mobile robot with Pioneer II by ActiveMedia, a laptop PC and two pan-tilt-zoom cameras.*

It can be seen that the greatest number of times occurred during transition relevance places (TPRs), or those moments in which the explanation of a concept is finished and the gaze is turned to understand if the concept has been assimilated or there are some questions. These results have been implemented in a prototype museum guide robot. The prototype (**Figure 1**) consists of two pan-tilt-zoom cameras (EVI-D100, SONY); a plastic head was attached to the top and the head movement was recreated by the pan-tilt mechanism of the camera. The lower camera has the task of observing and identifying the visitor and thanks to Toshiba's face recognition software library [7], it is able to understand if it is observed by someone.

The prototype was configured in two modes: *fixed mode* (display with the gaze fixed on the job and without head movements) and *proposed mode* (display with glances at pre-established moments for the visitor) and was tested on a sample of people visiting an exhibition. Quantitatively, it emerged that for a total of 7 times in which the robot turned to the participants during the explanation of the work of art, the participants turned to the prototype on average 1.6 times in the fixed mode, while in the proposed mode on average 4.1 times [8].

With the second experiment, an attempt has been made to understand if it is an actual consequence of the movement of the robot or a simple reflection. In this case, Robovie-R ver.2 (**Figure 2**) was used: a humanoid robot with improved head movements and three cameras on the chest to recognize the visitor who is looking at him. It is equipped with both autonomous and remote-control modes; the second is activated when, during the explanation, the visitor continues to look at the robot (the visitor's head is monitored by one of the two side cameras); as soon as the robot realizes it, it turns to the visitors and asks: "Do you have any questions?". Subsequently, the robot command is passed to the operator who can control it remotely while answering the visitor's questions (at the moment an AI response mode has not been implemented).

During this experiment, the following ways of displaying two posters of an exhibition were proposed: *random mode* (the movement of the head occurs at random moments during the explanation) and *proposed mode* (same as the previous experiment). Analyzing the movements of the sample during the experiment, two emerged: *nodding* and *mutual gaze*. The first is the vertical movement of the head towards the robot, a symbol of the fact that the visitor is understanding the exposure; if the movement occurs at a TRPs it serves as an invitation to continue the exposure. The second is a demonstration of commitment to understand. The combination of the two gestures, therefore, suggests understanding and involvement by the visitor towards the exhibition and the robot. From the **Figure 3**, it can be seen that the percentages between the two movements are close to each other in the proposed mode, this indicates that visitors are more involved and appreciate this type of interaction with robots more.

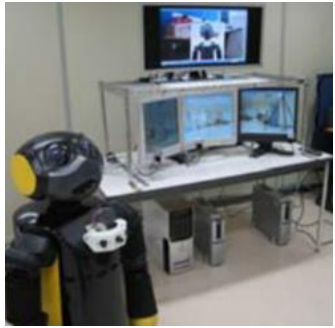

**Figure 2:** *Robovie-R ver.2 and the remote site.*

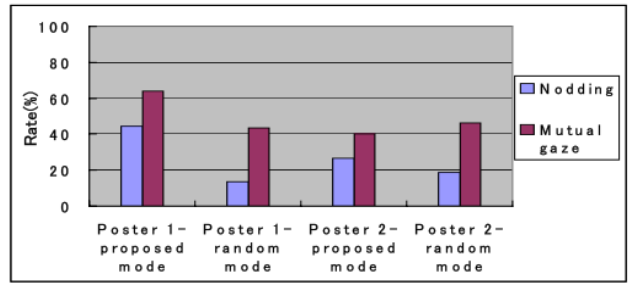

**Figure 3:** *Ratio of participants' reaction toward each poster in the two modes.*

### 2.1.2. RHINO

RHINO (**Figure 4**) is an autonomous mobile robot that has been developed to organize interactive tours for visitors to the "Deutsches Museum" museum in Bonn, Germany [6]. The objectives imposed at the beginning of the project were:

- *Safe and reliable navigation.* The robot must guide visitors, at walking pace, and not hit other people or the works on display. To achieve this objective, these additional problems must also be noted: RHINO must operate in very crowded places and often people can block movement sensors; although it is equipped with many types of sensor (laser, sonar, infrared and tactile), it is not guaranteed that it will detect "invisible" obstacles (for example, steel bars placed at different heights and glass cages for exhibits); finally, the layout of the museum often changes.

  - *Intuitive and appealing user interaction.* Since people who visit museums have a very variable age (2-80 years), the way the robot interacts with visitors must be "intuitive" and user-friendly.

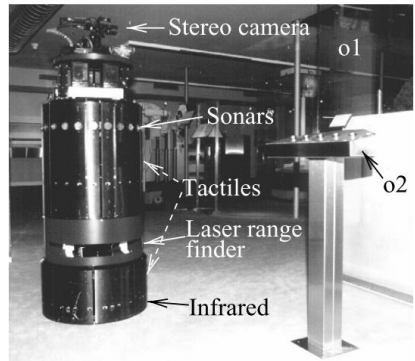

**Figure 4:** *RHINO and its sensors.*

RHINO was tested for 6 days inside the museum, for a total of 47 hours of activity in which it organized 2400 tours for 2000 visitors and more than 600 virtual visitors (the robot is equipped with cameras that also allow web users to participate in the tour). Only 6 tours weren't completed due to the change of batteries, so it achieved a success rate of 99.75%. He covered a total of 18.6 km with an average of 36.6 cm/sec (and a maximum of 80 cm/sec) and collided with obstacles 6 times without causing damage, of which only once due to a software failure (other times due to delays in battery change and errors in the museum map generated manually).

RHINO has been very successful in groups of all ages, but what were the software choices that led to such good results? The components linked to the two objectives described above play a fundamental role:

- *Navigation* includes two main functions: the first is perception which has the task of identifying the position of the robot (localization) and that of estimating the positions of the obstacles (mapping). The second is control with tasks related to real-time collision avoidance, path planning and mission planning.
    - The *location* of RHINO is based on a metric version of the *Markov localization* [9], estimates the position in an x-y-θ space (x and y identify the robot in a 2D Cartesian coordinate system, while θ the orientation). This approach uses probabilistic notions to reconstruct the position of the robot but is valid only in static and uncrowded environments. To stem the problem, it has been decided to implement an "*entropy filter*" which, applying it to all proximity measurements, allows us to establish which are correct and which are incorrect. The defect of this choice is the fact that the robot cannot restore its position once lost; for this reason, a small random part of the measurements made is saved [10].
    - The *mapping* of RHINO is greatly simplified thanks to the restricted domain; it is, in fact, enough to provide the robot with the map of the museum, even if it is necessary to update it often. The robot is equipped with a *probabilistic occupancy grid* [11] to modify the initial map: it constructs a 2D grid starting from the map and each cell is assigned an "occupancy value" which represents the probability that it contains an obstacle. By training a backpropagation network, occupancy maps can be created starting from sensor measurements

in the environment of interest. After training, the network generates probabilities of occupation [12].

- *Collision avoidance* prevents RHINO from crashing into works and/or people. An extension of the *dynamic window algorithm* (DWA) [13] is implemented which, thanks to hard and soft constraints, measures the robot's translational and rotational speed several times per second. The hard constraints limit the speed, while on the other hand, the soft ones increase the control. DWA only uses two soft constraints in order to allow RHINO to advance towards the target by maximizing speed; the basic algorithm, however, does not avoid "invisible" obstacles and for this reason, it must consult the map and obtain the coordinates of the obstacles. To obtain the coordinates, the maximum likelihood position estimate criterion calculated on the data arriving from the RHINO localization module was added to DWA. This algorithm is called μDWA [14] and, given the position of the robot, generates the "virtual" proximity measurements and integrates them with the real ones obtained by the sensors.

- *Path planning* has the task of developing the routes from one exposure to another, adapting to the constantly changing environment. The RHINO path planner is an any-time algorithm (provides results when needed so that the robot does not have to interrupt the journey) [15] and is based on *value iteration*, a dynamic programming/reinforcement learning algorithm [16], [17]. The algorithm works as follows: the value 0 is assigned to the cell containing the arrival, to all the others ∞. Next, calculate and update the values of each cell by adding the cost of moving up to the best value of the neighbouring unoccupied cells. After the convergence of all the values, the distance values of each cell from the finish are obtained. With the addition of two mechanisms, the processing efficiency is increased in real-time: the bounding box technique to focus the processing on the most important areas and an algorithm to recognize the areas to be recalculated when the map changes. The calculated paths are not executed unless after passing the collision avoidance routine.

- *Task planning* coordinates all RHINO actions related to interaction and movement. Furthermore, it has the task of transforming user-level commands into action sequences that the robot must do. The language used in this module is GOLOG [18], a language that allows you to translate complex actions into if-then-else and/or recursive procedures. GOLEX completes GOLOG by providing run-time components that convert the linear plans produced into hierarchical plans (sequences of elementary commands in machine language that the robot must perform) and conditional plans (sequences of elementary commands that are conditioned by the results of the detection operations) and which monitor their progress, thanks to time-out mechanisms and autonomous decision in case of lack of user input.

- The *user interface* is of fundamental importance for communication between man and machine, for this reason, it must be simple and accessible to all. RHINO has an interface for both "real" and "virtual" visitors.

    - The *on-board interface* (**Figure 5**) is an interface that collects different media (texts, explanations, graphics and sounds) useful for carrying out the possible tours. After listening to a brief explanation of how the robot works, the visitor

  interested in taking a tour of the museum can choose which one to go by pressing one of the 4 robot choice buttons. During the journey, RHINO shows a warning in his display when he intends to move to another work and indicates it by pointing with the camera placed on his head. Upon reaching a piece of work, he starts a brief explanation recording and waits for any requests from the user, if they are not there, continues the tour. He uses head movements to express his intentions (indicates the path he is following) and dissatisfactions (he uses his horn to get rid of the path by the people who block him). Once the tour is over, it automatically returns to the starting point.

o The *web interface* (**Figure 6**) is an interface that collects information web pages and Java applets for real-time updates on the movements and intentions of the robot. A RHINO web page is structured as follows: on the left are images taken by the robot, on the right from two rooms fixed to the wall, in the center the map of the museum and the point where it is located and, finally, at the bottom inside a box shows the actions carried out in real-time (in case the robot is explaining a job, both the description and the in-depth links are displayed).

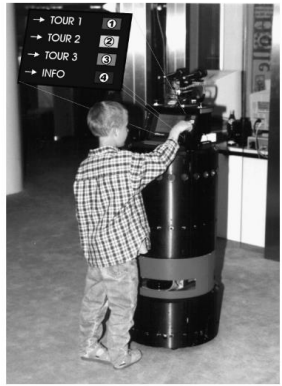

**Figure 5:** *RHINO's on-board user.*

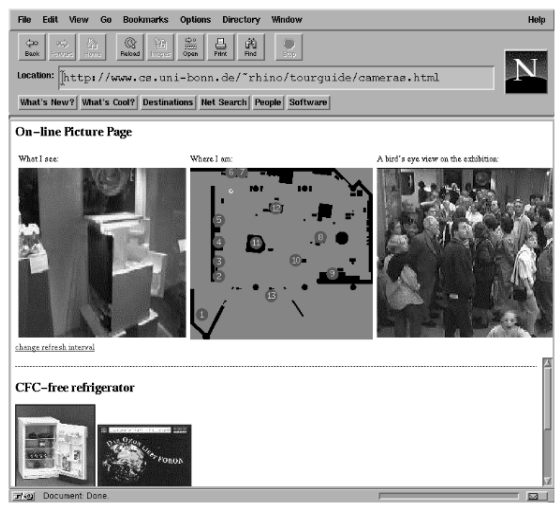

**Figure 6:** *An example of RHINO's web interface.*

## 2.2 Minerva, an algorithm of IA for the organization of exhibitions

Minerva [19]–[21] is a software tool that allows you to:

- automate the organization of the rooms of a museum for an exhibition
- offer the possibility to visit the entire installation "virtually".

Its architecture is very flexible and allows interaction with different types of users, changing the pattern of behavior. *Museum experts* and *museum organizer users* use the software to simulate potential exhibitions and evaluate their beauty (Minerva will be analyzed from this point of view because it is the most complete and interesting). *Artistic expert users* mainly request the display of the works shown, but do not affect the organization of the exhibition. *Inexperienced users* only carry out guided tours of already organized exhibitions. Finally, the *technical users* modify the system and update the instances of the works and environments databases.

But what is Minerva's architecture like? See **Figure 7**.

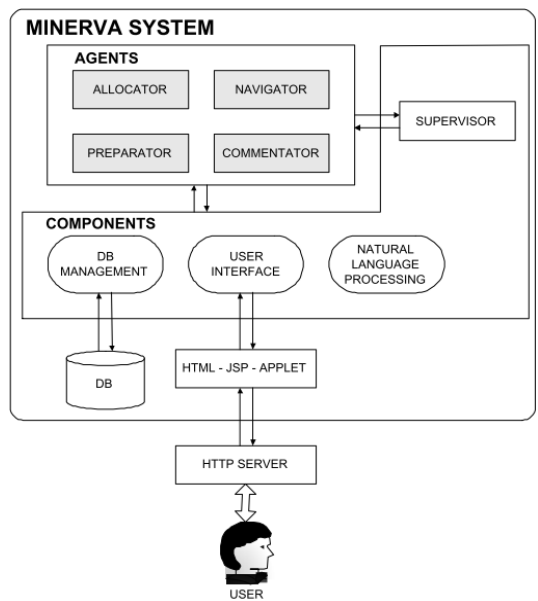

**Figure 7:** *The architecture of Minerva.*

Minerva is a multi-agent system developed in distributed artificial intelligence (DAI) and other components. The distinction between agents and components lies in the fact that the former carries out fundamental "intelligent" and autonomous activities, while the others deal with peripheral "service" activities. Its kernel consists of four agents, written in JAVA and they use the rule system called JESS [22] to carry out their inferential activities. The organization work is due to the collaboration between the various agents. They are:

- *Preparator agent* has the task of classifying the jobs and grouping them according to cultural and/or physical characteristics, all following the criteria provided by the user which are translated into JESS rules and inserted into the knowledge of the agent. Examples of criteria are the physical ones (some finds cannot be placed in rooms because with poor lighting or because they are too hot or cold) or artistic (groups of works are chosen that enhance the connections between them or to give emphasis to some works in particular). The result of the operation is stored in a taxonomic tree (**Figure 8**) in which each leaf corresponds to a group of works defined by the characteristics expressed by the nodes, which constitute the path from the root to the leaf itself. Each leaf is sorted according to its importance according to the criteria provided and, together with the selected environment, will constitute the input of the allocator agent.

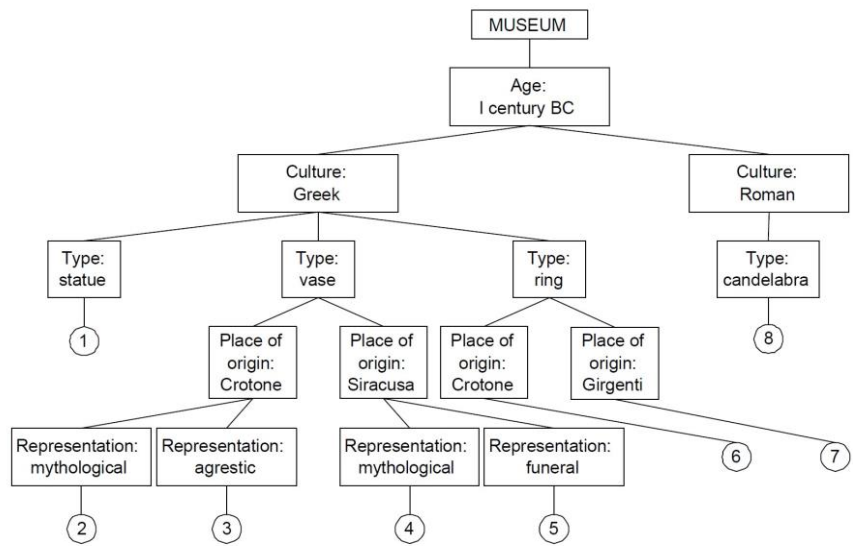

**Figure 8:** *An example of a taxonomic tree in which some finds of the first century BC are grouped.*

- *Allocator agent* undertakes to find the best arrangement of the groups of works provided (without modifying them) in the selected museum. Also, in this case, the criteria provided by the user (for example, the maximum and minimum occupancy of a room, the distance between one work and another and the free space that must be kept near the doors to facilitate the evacuation of the museum in case of danger) will be converted into rules written in JESS to enrich the knowledge of the agent. The groups are assigned to a room according to their importance, if the agent were to fail, the operation would repeat it by merging smaller groups into the same larger room or dividing a group into smaller rooms. Should all sorting attempts fail, the allocator agent asks the preparator agent to repeat the grouping operation. The result of the operation will be to describe how each object of the exhibition will be placed in the environment.

- *Navigator agent* (still in a primitive state of development) allows users who wish to visit the organized exhibition "virtually", creating routes according to the visit time set and providing an avatar in order to keep track of the progress made.

- *Commentator agent* (still in its primitive state of development) is activated when the user requests information about a piece of work during the tour; it creates a window in which information texts and three-dimensional models of the work being viewed are taken, taken from the collection database.

Other important components for the functioning of Minerva are:

- *Supervisor component* organizes and activates the components to perform a certain job.

- *User interface component*, generating HTML pages, deals with user interaction. The interface is in Italian because initially created for Italian users.

- *Natural language processing component* translates user requests from Italian into a format suitable for Minerva.

- *DB management component* translates queries into the format suitable for Minerva in SQL to query the databases of works and environments.

Provided a collection of works and an environment, Minerva can set everything up automatically or, since the user can adjust various parameters and set conditions on how to group objects, it can also be considered as a support and stimulus tool for user's creativity because Minerva takes care of carrying out all the less creative "low level" operations, which are nevertheless important to create an exhibition that respects all legislative and safety standards. In this way, more space can be left to the user's creativity in setting up. Minerva is not the only case of design support, but there are other software that can help human creativity in different fields, for example in the musical, literary, scientific and artistic fields.

The project is in continuous development to add functionality and improve the interface between the program and the user; currently, the latest "upgrades" that have been made to the software are improved interactivity with the user during the preparation process and JADE-based implementation. Another very important novelty is a second version of Minerva which places design objects, winners of the Compasso d'Oro award (a prestigious Italian design award), in a totally imaginary virtual environment. Furthermore, it is planned to apply the Minerva kernel to create software for organizing both real (for example the management of home automation devices in homes) and virtual objects (structuring of electronic courses).

## 2.3 Interactive exhibitions

An interactive and holographic AI-based system has been developed for exhibitions and cultural events to provide new ways of interacting and discovering Leonardo's art [23]. It has been used in both artistic and technological exhibitions in Milan, Naples and Genoa (**Figure 9**).

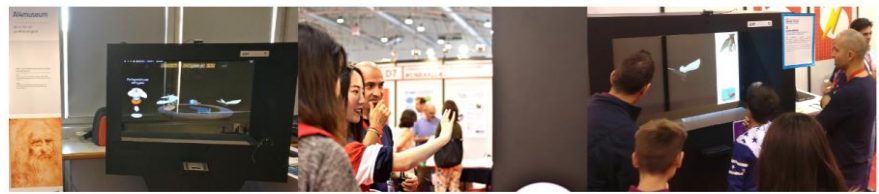

**Figure 9:** *Some photos of the system in artistic and technological exhibitions.*

The holographic effect of the system is based on the Pepper's Ghost technique [24]: the object to be displayed is projected onto a semi-transparent and semi-reflective glass arranged at 45° in respect to the source of the projection, in this way a holographic effect is recreated. The structure is 2 meters high and as a projection source it has a 55″ OLED TV; it also has several sensors: a webcam to detect the visitor, a microphone to record the voice and a Leap Motion controller under the projection area to manipulate the projected models.

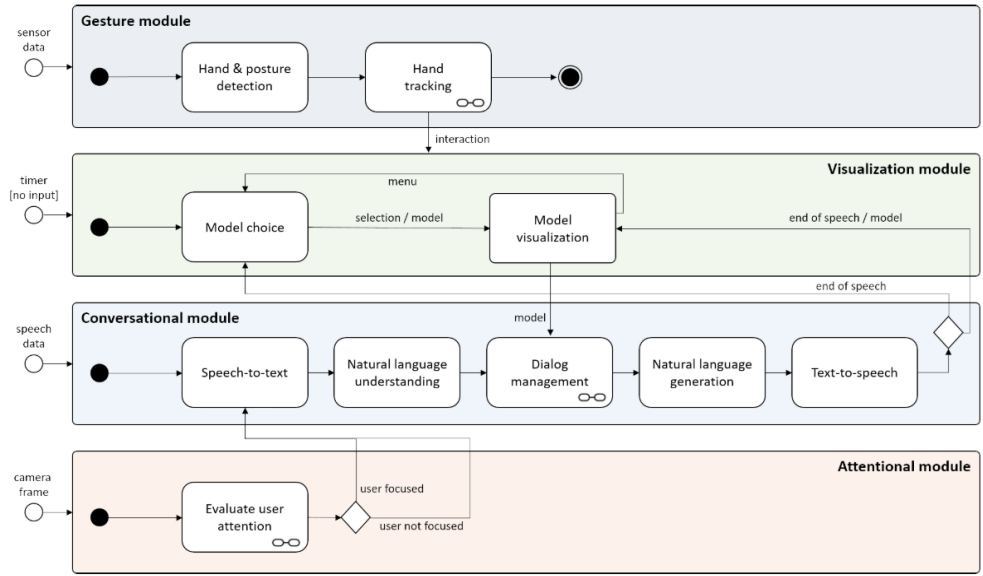

**Figure 10:** *Connections of the main modules of the system.*

The system (**Figure 10**) is divided into the following modules:

- *Gesture module* (**Figure 11**) is the module that detects the visitor's hand and acquires its position from the Leap Motion sensor. As visual feedback, it reproduces the user's hand and a wait-to-click metaphor is implemented as a selection tool. When viewing a model, two types of gestures are unlocked for manipulation: rotation and zooming. The system recognizes rotation if the visitor's hand moves left or right and begins to rotate the model in the desired direction; to stop the rotation, just place your hand in the center of the interaction area or remove it. The zoom is identified by the closing of the hand into a fist, the hand icon is replaced by a magnifying glass. The magnification level can be controlled by moving the hand in the sensor area forward (corresponds to increase zoom) and backward (corresponds to decrease zoom). To end the operation, the hand must be removed from the Leap Motion.

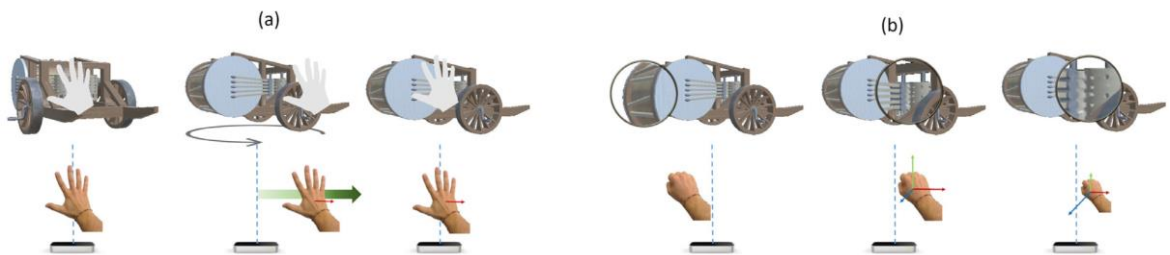

**Figure 11:** *How rotation (a) and zoom (b) techniques work.*

- *Visualization module* shows the visitor the entire catalogue of models that can be viewed by scrolling them clockwise and, if one is selected, takes care of displaying the chosen model. Thanks to the input of the Gesture module, he understands how to manipulate the 3D object. If the user asks a question, the *Conversational module* will communicate which part of the model must show during the response produced by the system. Return to the virtual carillon of the models, if the visitor has selected a virtual widget or if he does not receive input from the user after a certain set time.

- *Conversational module* is responsible for all the tasks related to communication with the user: understanding the request, establishing and managing a dialogue in natural language, asking for any missing information, extracting information from the knowledge of the system and processing the response to be provided to the user. Conversion from speech to text is possible thanks to Google Speech APIs, while the understanding of the text of the request is entrusted to the IBM Cloud Watson Natural Language Classifier (NLC). NLC was trained in a supervised way with a dataset of questions and, for each of them, the corresponding intent assigned by a man. NLC identifies the entities in the request and after looking for them in a dictionary built for the specific domain of interest, recognizes the type of request: *command intents* (model user command requests, the answers are predefined and disconnected from knowledge), *answer intents* ( model the user's answers to questions asked by the system), *factoid questions intents* (model the user's questions on the AI assistant or the cultural entities of the domain). System knowledge is organized in a Knowledge Graph (KG), saved as triple RDF and explored thanks to the SPARQL language; moreover, it is closed, that is, the answers are not extracted from the web, but only from closed and certified sources in order to prevent errors and incorrect information.

- *Attentional module* has the task of determining whether the visitor is talking to the system or doing something else (for example talking to someone else). it detects whether the visitor's gaze is turned to the hologram or not and, depending on the outcome, activates or disables the *Conversational module*.

## 3. Algorithms of artificial intelligence for the reconstruction of finds

### 3.1 Reconstruction of pottery

The classification of ceramics can be defined as the practice of describing and arranging the artefacts found in taxonomic groups, observing and comparing different characteristics of the same. Artefacts are classified based on information:

- *Structural* refers to the geometric structure

- *Mereological* belongs to the relationship between different parts that make up the artefact

- *Morphological* is related to the form

- *Functionals* are related to the function of the artefact

- *Materials* refer to the chemical-physical composition and the material with which the objects are made.

The research, in this context of study, has as main objective the identification of a regular or irregular trend in the spatial-temporal distribution of certain corpus of materials, in order to detect useful elements for the identification and understanding of moments of continuity, contact, transmission or transformation that characterized the societies of the past; therefore the artefacts found can be observed as the main witnesses of the final result of production processes, on the one hand, and practical and ideological-symbolic uses, on the other. The classification of the ceramics not only provides a systematic description of the material found but also aims to reconstruct fragments in whole forms or to compare different fragments stored within a database. These operations are quite laborious and require a lot of time and great knowledge of the sector, for this reason, different automatic or semi-automatic methods are being experimented to carry them out [25], [26]. The approaches on which the automated or semi-automated methods for classification and topological attribution are based are:

- *Sub-symbolic approach* is based on the use of neuronal architecture, that is, a multitude of interconnected elementary logical units that do not need a priori knowledge to carry out their work: these networks, called "neural networks", have the ability to learn from experience, without any mechanism that determines its starting behavior. To train it, it "trains" by providing it with a collection of data (both positive and negative), each logical unit (neurons) calculates with a simple calculation a binary result on the basis of the numerical values (weights) that are at its input channels. The learning of the network takes place when the weights relating to the logical unit of which it is composed are updated in order to provide relevant answers to the problem. The algorithm used for this operation is "backpropagation error", which adjusts the weights of its logical units in order to reduce as much as possible the distance between the actual response and that assigned by the instructor.

- *Symbolic approach* consists in representing, by means of a suitable formal language

(machine-readable), the knowledge that a given agent possesses, relative to a given domain of interest, in a "knowledge base" and from this base draw the necessary conclusions the resolution of a specific problem by means of inferential manipulation mechanisms of the symbols present in the base. It has a lot in common with logic and mathematical theories: knowledge of a given piece of the world is represented in the form of axioms from which conclusions are drawn (theorems).

- *Data mining* characterized by the application of statistical numerical analysis algorithms and usually dedicated to solving clustering problems. It should be noted that for clustering the data is grouped into groups by the algorithm, to which the a priori classes are not provided which, instead, occurs in the classification.

To support the typological classification of ceramic artifacts, current literature shows an increasing use of methods based on neural networks such as, for example, methods that use ultrasound techniques and advanced pattern recognition techniques [27], others that study the characteristics of the curves of the 3D models of archeological artifacts [28] … Two noteworthy projects will be exhibited: the ArchAIDE project and a project for the recognition of ostraka fragments.

### 3.1.1 The project ArchAIDE, Archaeological Automatic Interpretation and Documentation of cEramics

The ArchAIDE project [29]–[31] is a three-year European project (June 2016 - May 2019) and was coordinated by the Department of Civilization and Form of Knowledge of the University of Pisa. The consortium included research institutions and small and medium-sized enterprises from five different countries: Italy, Spain, the United Kingdom, Germany and Israel.

The idea behind ArchAIDE is to support the classification and interpretation work of the archeologist with innovative IT tools for the automatic recognition of archeological ceramics, through the development of a software platform available on mobile devices (tablets, smartphones) and on desktop computer. The classification of ceramics is of fundamental importance for understanding and dating archeological contexts and for understanding the dynamics of production, commercial flows and commercial interactions. The digital automation of this process could revolutionize archeological practice, reducing time and costs, improving access and enhancement of the digital cultural heritage in a sustainable way and allowing a deeper knowledge of archeological contexts. Furthermore, it is designed to allow access to tools and services that can improve the digital archeological resources available, through the possibility of publishing the results of the classification as open data, and tools for analyzing and displaying data, providing useful tools for archeological interpretation. To grasp the similarities between the ceramic products, the conditions of installation, lighting and resolution and the specific characteristics of the ceramic must be unchanged. These conditions can be treated using deep learning algorithms that analyze the shape and the decorative aspect (elements that contribute most to the classification of the ceramics), while the similarity based on the appearance is obtained using a Convolutional Neural Network (CNN) [32].

In recent years, many scholars from different disciplines, including archaeology, have begun to wonder whether a Big Data approach can be applied from both a theoretical and practical point of view. Big Data means working with a set containing all possible data (or almost) from different disciplines and which can be useful in solving a problem [33]; this approach allows you to have a global view of the data: you can explore from different angles, observe common characteristics that would not be observable with small amounts of data. The exponential increase in the data that can be processed is due to the phenomenon of digitization, or the translation of physical information into a computer-readable language. This, however, does not imply datafication: the transformation of actions or processes etc. in a quantized format that can be tabulated and analyzed [34]. While digitization uses automatic sampling methods, datafication adapts to a Big Data approach and techniques that allow you to analyze nonlinear relationships between data. The datafication allows to go beyond the simple digitization by making available new opportunities for non-linear data analysis, in addition to the traditional ones: data flows could be obtained starting from the data collected in the field derived from the positions and relationships between the finds and sites.

ArchAIDE's work is configured as *proof of concept*, whose purpose is to demonstrate that the methodology and technology used can recognize the shapes and decorations of the ceramics starting from the photographs taken by archeologists in the field. The key characteristics selected to define a ceramic form are seven and they are: external profile, internal of the shape, external profile, internal of the sockets, section of the sockets, maximum height point (rim point), maximum point of the base (base point); these measures have been associated with a scale factor to be able to compare them in reality. To unify the diversity of the various catalogues, it was decided to create two elements in the database:

- *Reference database* contains all the data on the definitions of the types of ceramics, decorations, stamps and mixtures and those necessary for the analysis of the fragments; its main entities are:

    o *Pottery type* is the entity that represents the descriptions of the individual types (forms) used for the recognition of fragments with attached information on the places of production and discovery and attached to images, bibliographic references, etc.;

    o *Decoration* is the entity that collects descriptions of decorative genres with information on colors, chronology, places of production and discovery and attached images and bibliographic references;

    o *Stamps* are the entities that contain the description of the stamps/brands with which the shops/potters signed their products, as well as descriptive information and multimedia files.

- *Results database* contains data on the fragments collected in the field.

Managing all the data in the same context allows you to create a unique dataset for use in data mining and visualization and to avoid potential inconsistencies resulting from the separation of two datasets that share much of the information. In order to populate the database, it was decided to convert the paper catalogues into digital descriptions: the drawings and textual parts were extracted from the scan. For the extraction of the text from the scan, an open-source tool in Javascript OCR called Tesseract OCR [35] was used. Once extracted and organized, all the information is archived in an annotated SVG file with semantic information. Another part of the database comes from other already existing sources provided by the project partners. Data migration took place using JSON XML or SQL queries directly. In the future, the possibility of catalogues that require semi-manual or even fully manual digitization cannot be excluded.

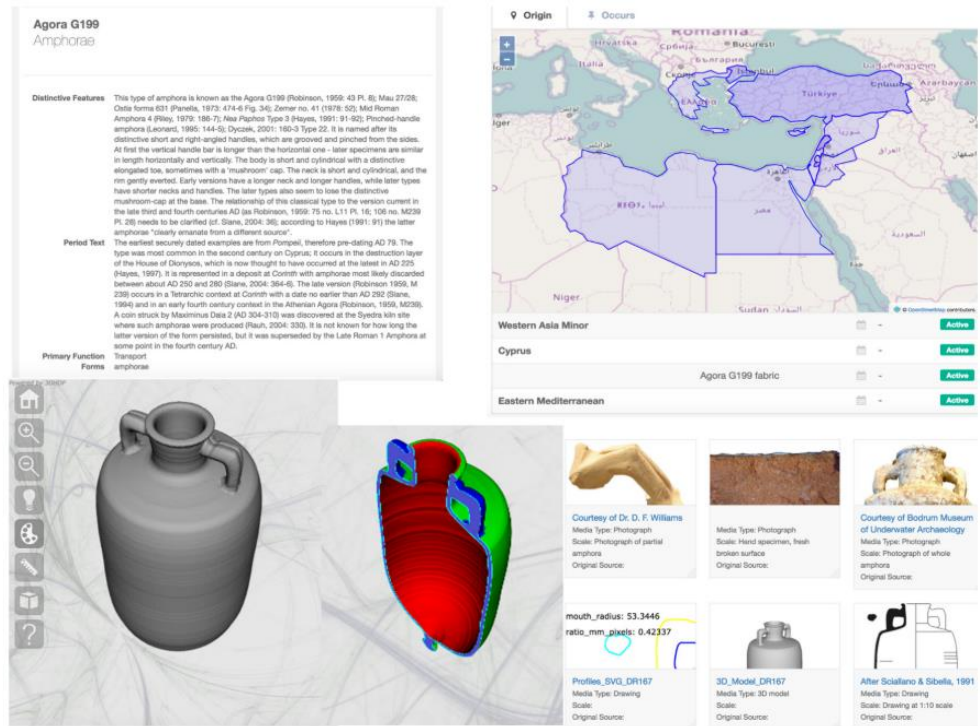

**Figure 12:** *The information displayed in the database: textual part, information on the place/areas of discovery, multimedia information and interactive navigation through the 3DHOP platform of the 3D model of the shape.*

The database page (**Figure 12**) associated with each ceramic type contains: the descriptive textual part, information on the locations/areas of discovery, images, SVG of the extracted profile and navigation through the 3DHOP platform [36] of the 3D model of the shape.

ArchAIDE has developed a recognition system based on a Deep Learning approach, which allows both *shape-based recognition* and *appearance-based recognition* starting from a photograph taken by the user via a mobile device.

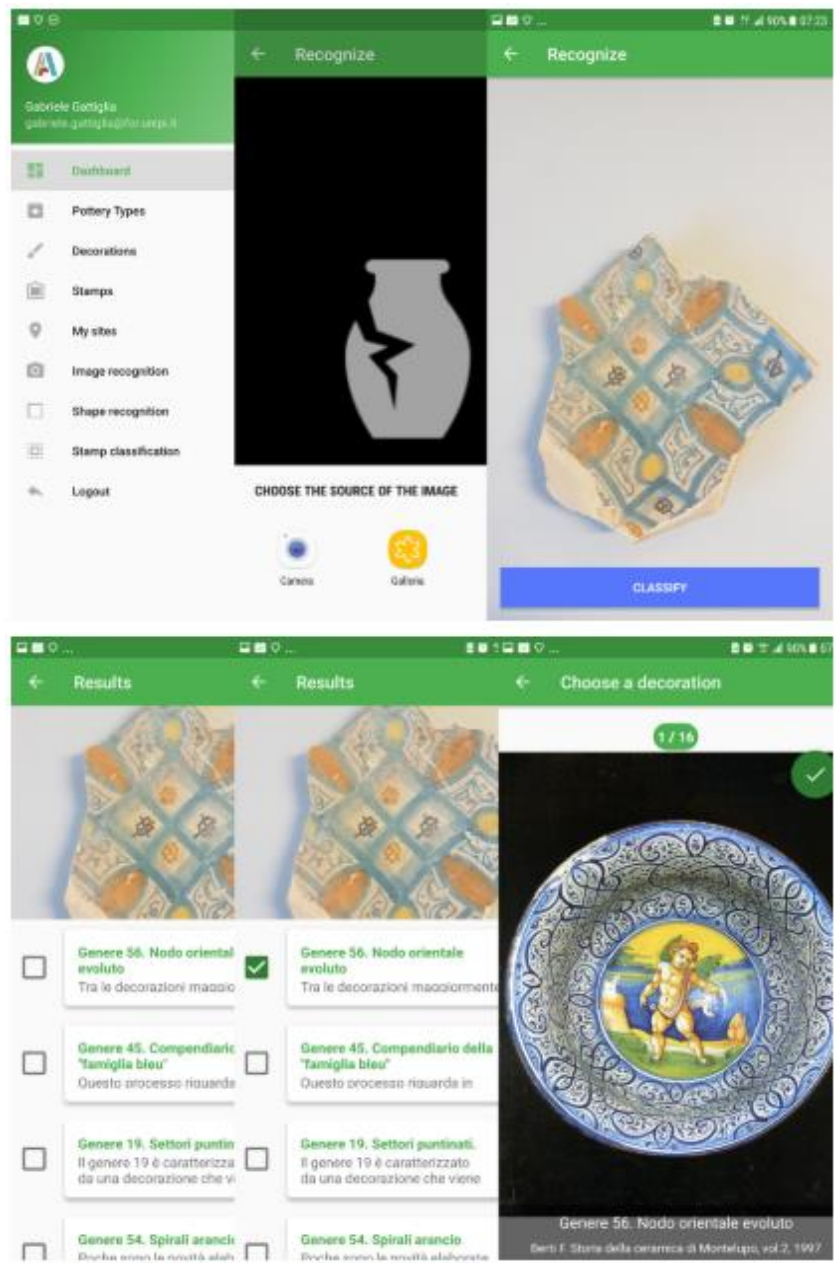

**Figure 13:** *Screenshots of the ArchAIDE app available on the Play Store. After taking a photo of the fragment to be recognized, recognition is activated via the Classify option. The system returns the 5 most probable results with relative percentage, the user can validate the answer by comparing the fragment with the database images.*

To instruct the system, a correct definition of the classification and a very large training set are required: to fill the lack of data for the *shape-based recognition*, in addition to the data of real examples, it has been chosen to fragment the 3D models to generate a dataset large enough to be able to perform a training and testing suitable to achieve the best possible system accuracy. The input, in the form-based classification system, is represented by SVG files containing the characteristics of both the types and fragments. The classification of a partial shape can be difficult, since the number of combinations of alignment between two shapes is practically infinite; in addition, similarity measures should also be taken into account. On the other hand, to increase the quantity of examples to be used for the classification of ceramic decoration, a hybrid solution was chosen: multiple images were created starting from the same thanks to cutouts, stairs and overturnings and given to a neural network pre-trained to classify images into categories outside the archeological sphere. All of the above is the result of the first year of work on the ArchAIDE project, subsequently the automatic recognition system will be implemented and tested on mobile devices and mobile platforms (**Figure 13**).

In 2018, the appearance-based recognition functionality was added to the ArchAIDE app to be tested by field scholars and all the awards made will be added to the database and shared, integrated with information on cultural heritage that come from different sources and can be displayed in order to produce a truly significant impact in the progress of the discipline and in accessibility for professional and non-professional users.

### 3.1.2 Pairing and reconstruction of ostraka with neural networks

With the word *ostraka*, the ancient Greeks indicated the tortoiseshell, the shells and other objects of hollow and round shape such as terracotta pots and in particular the fragments of such pottery (**Figure 14**). By convention this name is still given to the shards, often used in antiquity as an extremely inexpensive writing material; many have been found in Upper Egypt and Numidia dating from the Greek-Roman period to the Arab period and, very often, contain Egyptian inscriptions (mainly in demotic, some also in hieratic and hieroglyphic).

The study [37] was carried out starting from a dataset consisting of ostraka with demotic inscriptions, originally written on discarded ceramic fragments; since the fragments are mainly flat and it has been decided to study their engraved content, it has been chosen to focus on 2D reconstruction techniques by approximating them to "documents". Reconstruction of images from fragments has been used in many contexts (for example, automatic puzzle solving); only in these years, clustering techniques have been designed to group fragments of the same type and then recreate each image by ordering the pieces identified previously. A more sophisticated variant of this problem is the reconstruction of ancient artefacts: due to the conservation conditions, the fragments can have ruined and worn edges or, in the worst case, be absent [38].

The strategy used to solve the problem consists in the use of a Convolutional Neural Network (CNN) [32], or a particular type of multi-layer architecture of Deep Neural Network (DNN). To compare two images, an architecture called Siamese Neural Network was used: it consists of two identical branches, which share the same weights during all phases. This type of net is trained to distinguish similar pairs from different ones, it can be used for facial recognition or for one-shot alphabet recognition.

The ostraka ceramics have small variations in color and inscriptions, therefore, before focusing on the overall assembly, similarity has been sought between pairs of fragments. The starting dataset is made up of 2D RGB images in high resolution (greater than 3000×2000px) of 30 large fragments. It was decided to divide them into non-overlapping pieces of 400×400px, introducing "artificial fractures": neighboring fragments are joined and then cut along a diagonal that crosses the seam and cut out so that they maintain the same starting size. Furthermore, a label has been assigned to each type of matching: if two fragments are not close, the value 0 is assigned, if they are two consecutive, a value between 1 and 4 is given according to their mutual position (1 = right, 2 = left, 3 = above, 4 = below).

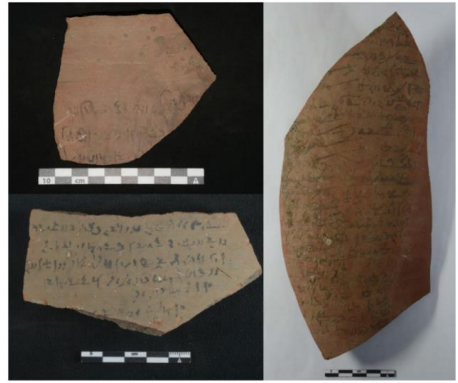

**Figure 14:** *Examples of ostraka with the scale bar below.*

The architecture of the proposed network (**Figure 15**) is a *2D Siamese Convolutional Network*, with two branches born from the succession of "Conv Block"; each block consists of: a 2D convolutional layer, a batch normalization layer, a leaky ReLu activation layer and a 2D max-pooling layer. To limit the calculations made in the four junction areas of each fragment and the use of network memory, the "Crop" layers cut bands of 10px from the 4 edges of the image. In the "Subtract" level, "change-maps" are created thanks to the subtractions between the two sub-feature maps leaving the two branches of the network. The absolute value of these subtractions is passed to the next level, therefore it is ensured that the model is concerned with calculating the difference only in the four joint areas. The change maps are then chained and sent to the "Global Average Pooling" (GAP). The last part of the network consists of a dense layer, a dropout layer and finally a dense layer with softmax activation to calculate the output (a number between 0 and 4, as explained above). The entire structure is built using the Tensorflow [39] and Keras [40] frameworks, suitable for deep learning. This exposed network reached an accuracy of 81% on a small dataset (900 fragments) and 96% using a larger dataset (7000 fragments).

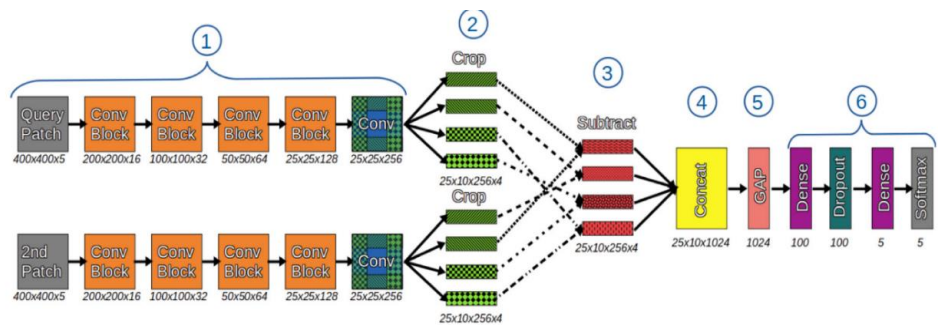

**Figure 15:** *Simplified architecture of this project. 1) Siamese CNN architecture, 2) ROI cropping in each direction, 3) Absolute value of the subtraction between ROIs from query patch to corresponding ROIs from 2nd patch, 4) Concatenation of intermediate results for each alignment direction, 5) Global Average Pooling, 6) Classification: The output is a vector of probabilities for each of the five classes.*

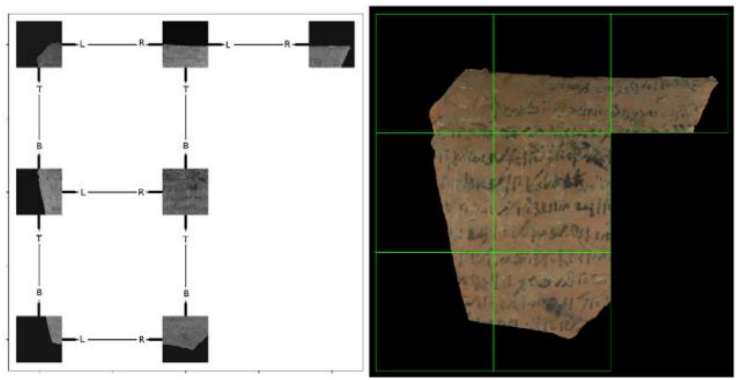

**Figure 16:** *An example of global reconstruction with its graph: each edge keeps track of the spatial arrangement of the fragments with concerning the others.*

The second part of the project aims at the global reconstruction of the ceramics: the elaborated method is based on the construction of a direct graph in which the nodes are pieces of image that can have at most four neighbors (1 per direction). Each branch leaving the node contains the value resulting from the network and therefore the probability that the two fragments are consecutive. Two fragments are considered consecutive if they get a probability greater than 0.7 (**Figure 16**). The method offers various possibilities of reconstruction and, sometimes due to unexpected alignments, even with missing fragments; it is then up to the scholar to decide which are the most realistic. This problem is due to the insufficiency of ceramic texture variations, on the contrary the incisions, stains and scratches on the surface increase the ease of reconstruction.

In future work, efforts will be made to reduce incorrect alignments and provide an interactive interface that allows archaeologists to manually select the best reconstruction proposal.

## 3.2 Reconstruction of texts and epigraphs

*Epigraphy* is one of the main sources of knowledge of the uses and customs of ancient civilizations and studies the inscriptions on poorly corruptible materials; its dominion is very vast and varies according to the historical context: it passes from tombstones and funerary monuments to commonly used artefacts and weapons (**Figure 17**). Since most of the inscriptions are damaged by time and arrive fragmentary and illegible to the present day, epigraphists study the context in which they are made and, subsequently, hypothesize possible reconstructions of the original.

In the following study [41], a method was developed to automate and speed up this procedure. It has been chosen to use the epigraphy of ancient Greece because it is very vast and variable, which makes it an excellent challenge for Natural Language Processing (NLP) and because large digital collections of ancient Greek texts have recently been created (PHI [42], First1KGreek [43] and Perseus [44]).

In training, a modified version of PHI was used (the largest dataset containing inscriptions from ancient Greek epigraphs), called PHI-LM. In PHI, scholars, by convention, mark the missing characters with dashes, convert the text to lowercase and add punctuation marks and diacritical marks (*Leiden Convention*); all these changes are also found in PHI-LM, but since the scholars' notations were often inconsistent and noisy, it was decided to write a pipeline to convert the text and make it usable by the machine. After calculating the frequency and standardizing the alphabet used in the inscriptions, including the accented letters (147 characters), numbers, spaces and punctuation marks, a new symbology is introduced: the '-' represent a missing character, while the '?' indicates a character to interpret. Then the texts of the collection were rewritten by eliminating all the elements that could "confuse" the network: annotations and comments not in Greek, any spacing and punctuation errors. Furthermore, the number of missing characters has been associated with those assumed by the epigraphists, thus converting the length value into an equal number of dashes. Finally, texts with less than 100 characters have been removed. In this way PHI-ML was created; the dataset contains more than 3.2 million words and is used to train PYTHIA.

The architecture of PYTHIA (**Figure 18**) is a sequence-to-sequence [45] based on the architecture of the neural network: it consists of a Long-Short Term Memory (LSTM) [46] encoder, an LSTM decoder and an attention mechanism [47]. The encoder takes the entry text in input and passes it through a lookup table with learnable embedding vectors. Subsequently, the input passes into the decoder which is responsible for predicting the content of the '?'.

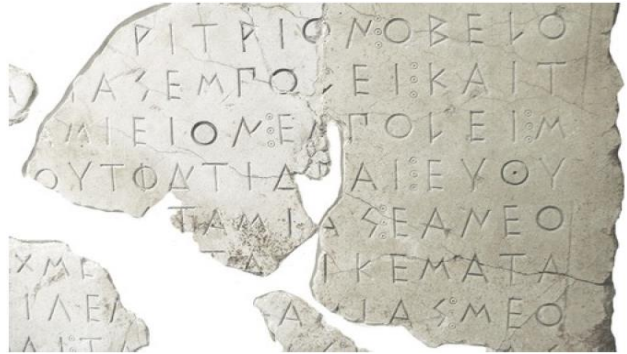

**Figure 17:** *An example of an inscription damaged by time found in the acropolis of Athens and dating back to 485/4 BC.*

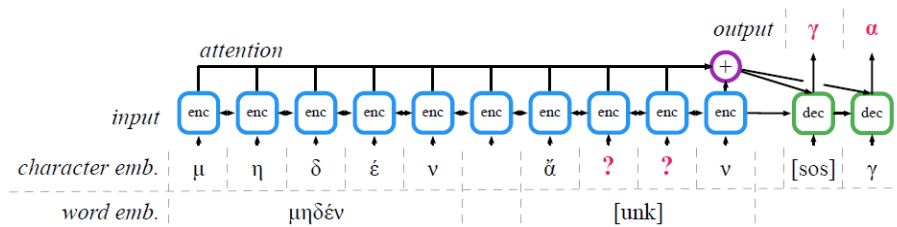

**Figure 18:** *An example of the functioning of PYTHIA-BI-WORD: in this case the input phrase is μηδέν ἄγαν (medénágan) "nothing in excess", a maximum inscribed in the temple of Apollo in Delphi. The characters to be predicted ("γα") are marked with '?'. When an incomplete word enters the network, it is treated as unknown, the decoder tries to reconstruct the correct word and, finally, it takes care of returning the output.*

**Figure 19:** *An example of reconstruction of an inscription; the areas highlighted in blue are correctly deduced, the purple ones incorrect.*

To improve the performance of the system and focus on the meaning of the words rather than on identifying the characters, it was decided to make sure that the encoder can receive an additional input stream and the decoder focuses only on the current output. To facilitate the recognition of words by the system, a list has been generated with the 100,000 most frequent words in the dataset; this table together with a separate lookup table can be concatenated, step by step, to incorporate each character with the word to which it belongs. Words not contained in the table or with missing characters are mapped as unknown ('unk'). To assess the effectiveness of PHYTIA, parts of the inscriptions in input of varying length (between 100 and 1000 words) have been removed and replaced with special prediction characters. The sequences of characters to be deducted are between 1 and 10 consecutives characters.

The **Table 3** shows the results obtained and compares them with other approaches; the first value (CER) indicates the character error rate, while the second provides the probability that, in the top 20 of the possible translations, the true reconstruction of the inscription is present. Notice that PHYTIA obtains a CER lower that that obtained by PhD students.

| Method | CER | Top 20 |
|---|---|---|
| **Ancient Historian:** Two PhD students with epigraphic competence. | 57,3% | ----- |
| **LM Philology:** Standard count-based n-gram language model (LM), trained on and then tested on PHI-ML. | 68,1% | 26,0% |
| **LM Philology & Epigraphy:** Language model trained on First1KGreek, Perseus and PHI-ML. | 65,0% | 28,8% |
| **LM Epigraphy:** Language model trained on PHI-ML. | 52,7% | 47,0% |
| **PYTHIA-UNI:** architecture that accepts only input characters and has unidirectional LSTM. | 42,2% | 60,6% |
| **PYTHIA-BI:** architecture similar to the previous one, but with bidirectional LSTM. | 32,5% | 71,1% |
| **PYTHIA-BI-WORD** | 30,1% | 73,5% |

**Table 3:** *It compares the CER and Top-20 values between epigraphists, language model variants (LM) and variants of PYTHIA, the study network.*

Since PHYTIA is the first model of reconstruction of ancient texts of this type and in the hope that it will be of help to future research and inspiration for future interdisciplinary works, it has been chosen to publish it as open source (**Figure 19**).

## 4. Algorithms of artificial intelligence for the study of human remains

The careful observation of the shape of the human body (as well as its faithful reproduction) has its roots in the Greek artistic culture of the classical period up to the erudite tradition of the Latin world, but it developed above all during the Renaissance. Leonardo da Vinci first resumed the study of the skeleton and the human body to search for the parameters that defined the proportions of the human figure and came to formulate new anthropometric principles [48]. He was responsible for the first exact representation of the skeleton and the merit of having developed the first scientific anatomical iconography.

In the mid-17th century, the progressive emancipation of science and the affirmation of the experimental method in the scientific environment led to an interest in the naturalistic observation of man. Anthropometry was born thanks to the development of anatomical knowledge and the possibility of observing body variations and proportions. During the second half of the 19th century, the interest in these studies extended to the bone finds of ancient populations found in archeological excavations, in which only the skulls were collected, to determine the properties that could define and distinguish the individuals and populations in "types".

From the middle of the 19th century, thanks to the rediscovery of the laws of inheritance by G. Mendel and the definition of the genetics of populations, a radical conceptual and methodological renewal process began in the way of tackling the study of human diversity: the variability and history of human groups can be understood as a result of their biological history and adaptation to different natural environments. The biological (hereditary and non-hereditary) characteristics of a human group overlap and interact with the cultural component - understood as a system of social, economic, behavioral relationships, etc. proper to each community - in regulating the activities and relationships that individuals can express. Current research is aimed at giving the study of human remains a growing functional connotation, which can make it possible to acquire information on the dynamic processes connected with the life of human groups of the past, such as the demographic structure of the living group, the state of health and disease, the nutritional situation, the existence of kinship relationships between groups of individuals and, last but not least, the population of entire geographical areas.

In recent years, efforts have been made to create artificial intelligence-based tools to simplify and automate these anthropological studies. In this session, two studies on human remains are analyzed, namely the estimate of the height starting from the bones and the determination of the sex starting from the skull.

### 4.1 Determination of sex from human skulls

Current methods of determining sex from human remains are based primarily on morphological and metric criteria. These techniques are not only complicated and time-consuming but also very subjective and great experience is required. Studies of these methods have reported different accuracy; in addition, in physical and forensic anthropology there is much debate about which are the best discriminators for determining sex between measurements of the anatomy of the skeleton and morphological characteristics.

In the study shown [49] below, a neural network is proposed that is trained to recognize, starting from photos of skulls, the sex of the individual. Neural networks are very useful for this task: thanks to their structure at levels of connected "neurons" which each analyze a different part of the image and finally the output probability of belonging to a class of the input image is processed. The use of artificial intelligence techniques circumvents the problem explained above because, depending on the case, it will give more weight to the most useful components for the determination, discarding those that are not useful.

For the neural network to learn to recognize sexual dysmorphias from skulls, a large number of examples with known sex are needed: 3D reconstructions of skulls based on CT scans of the Royal Adelaide Hospital were used in this study. The dataset is made up of 500 CT studies of male and female skulls aged between 18 and 60 and of different ethnicities (although the prevalence is European). All scans were performed on Siemens SOMATOM or Toshiba Scanner AQUILION CT and, after removing all patient identification information, each of them was post-processed on a standard radiology station with PACS Carestream (Carestream Health, Rochester NY). Thanks to a built-in function of the PACS Carestream it was possible to create, for each case, a manipulable 3D skull from which 2D images were taken in jpg format. The images taken of the skull are of the left side with a slight variability in orientation, not all of them have facial bones and/or the upper part of the cervical spine (they are not dysmorphic elements such as the skull, therefore they do not affect the result).

It has been decided to use the *transfer learning technique*, that is to modify an already existing neural network to solve a problem. The network used is GoogLeNet, a *Convolutional Neural Network* (CNN) [32] trained to recognize 1000 commonly used objects in images [50]. Thanks to MATLAB (a programming environment written in C and from the programming language of the same name created by MathWorks used mainly for numerical calculation and statistical analysis, which also offers specific toolboxes for the use of machine learning, neural networks etc.) it was possible to modify the last 3 levels of the network to be able to classify, starting from the input images, if the image belongs to the "Male" or "Female" class [51]. Each convolutional level of a neural network works just like a brain neuron: it is activated in the presence of some peculiarities of the image, for example, a large protuberance of the occipital or the curvature of the glabella. It should be noted that the nets were not provided with a preliminary knowledge of the skull, but they learned using the backpropagation from the dataset provided during the training phase. Backpropagation allows the network to compare the estimated result with that provided and update the weights of the internal connections of the layers to obtain better results. The images to be supplied as input (**Figure 20**) must be 224×224 pixels in size and must first be subjected to image augmentation, a technique that introduces "random noise" in order to increase the variety of cases to avoid overfitting: the images were randomly rotated, translated and inverted in relation to the axes.

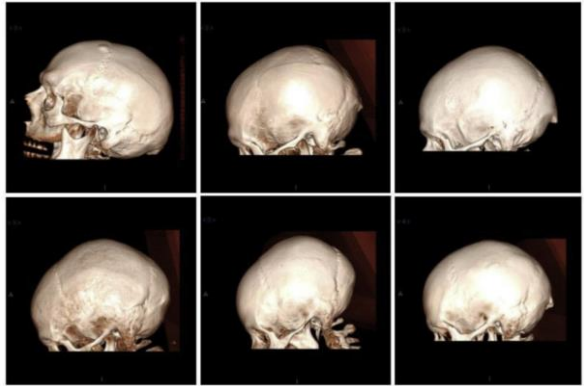

**Figure 20:** *examples of 3D reconstructions of skulls used as inputs (the top three are male skulls, the bottom three are female skulls).*

The final dataset obtained for training and testing the neural network has 1000 images and has been divided into 900 images for training and 100 for testing (both divisions have the same number of male and female cases). For 2000 epochs the system has been trained with 900 images divided into mini-batches of 90, at the end of the training the network is subjected to the test of 100 images never shown before and, based on the results obtained, the general accuracy of the system is calculated. The **Table 4** is a confusion matrix that shows the results distinguishing the correct estimates from the evaluation errors.

|  |  | Sex (Medical Record) | |
|---|---|---|---|
|  |  | **Male** | **Female** |
| **Sex (AI Estimation)** | **Male** | 47 | 2 |
|  | **Female** | 3 | 48 |

**Table 4:** *Shows the sex estimate results of the AI algorithm compared with medical records.*

To evaluate the recognition performance, it is also necessary to calculate the precision for each category: it is given by the ratio between the elements recognized correctly for that category (*true positive*) and all those recognized for the category, also recognized incorrectly (*false positive*):

$$Precision = \frac{tp}{tp + fp} \quad (1)$$

The precision for the "Male" class is equal to 0.96, while for the "Female" class it is 0.94. The general accuracy, however, is calculated as the ratio between all cases recognized correctly for each category and the total number of cases supplied to the network:

$$Overall\ Accurancy = \frac{\sum_{i=1}^{k} tp_i}{n} \quad (2)$$

where $k$ is the number of classes into which the cases are divided and $n$ the total number of the data set for the test. In the study, an accuracy of 0.95 was found. From the results that emerged, it can be said that artificial intelligence methods are a valid solution for the determination of sex starting from skulls because they do not need particular skills to be implemented, they are quick and eliminate human prejudices from the estimation process.

Future changes to this study could allow the use of data from the archeological environment and can also be applied to other parts of the skeleton with dysmorphic characters such as, for example, the pelvis. Furthermore, the source of the inputs to the network could be changed: not only the CT scans but also images taken by the camera or smartphone in order to offer rapid use in the field.

## 4.2 Prediction of stature starting from bone measurements

From an archeological point of view, understanding the stature of an individual starting from his remains is of fundamental importance because, together with age and weight, it allows researchers to evaluate sexual dysmorphism or the body size of the population analyzed. The stature of an individual provides a lot of information because it does not derive only from the natural genetic potential, but also from the society and the environment in which man develops. There are numerous approaches, more or less precise, to estimate the height starting from the human skeleton:

- *Anatomical methods* provide for the reconstruction of the skeleton by adding all the bones included from the skull to the heel and adding a certain height to the now absent soft tissues. These methods are more precise than the following, but not all bones are said to be available. The first anatomical method was proposed by Dwight in 1884 [52], although it led to a high degree of error.

- *Mathematical methods* are effective methods when there are few bones available and they are very fast, they imply regression formulas based on the correlation of specific bones and living stature. The first regression formula was introduced by Pearson in 1899 [53].

In [54] the auhors tried to estimate the stature thanks to two supervised regression models: one based on an Artificial Neural Network and the other based on Genetic Algorithms. Before proceeding, it is necessary to define what an *Artificial Neural Network* (ANN) (**Figure 21**) and a *Genetic Algorithm* (GA) (**Figure 22**) are.

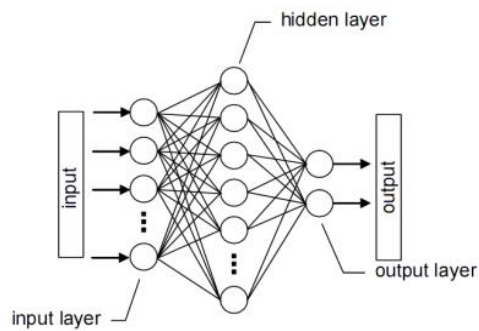
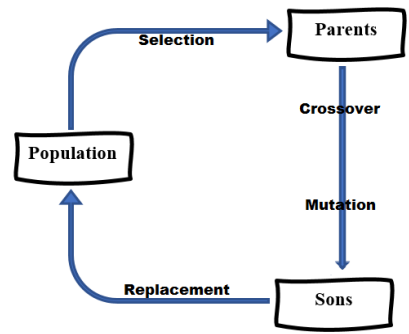

**Figure 21:** *An example of the architecture of an Artificial Neural Network (ANN).*

**Figure 22:** *Scheme of the functioning of Genetic Algorithm (GA).*

ANNs [55] are inspired by the human brain: they are made up of various units, called *neurons*, all interconnected with each other. Each *neuron* receives real values at the input and the product will become input for the following. It is an adaptive system because it allows you to adjust its internal parameters during the training phase and, subsequently, during the test you can measure the error between the output produced and the desired one and recalibrate the entire system.

GAs [56], on the other hand, are algorithms that are mainly used to solve research and optimization problems. Their functioning is very reminiscent of natural selection and evolution: only individuals best suited to changing the environment survive. The algorithm begins with a list of possible random generated solutions (*population*), each solution (*individual*) is associated with a value that is calculated by the fitness function and determines how excellent that solution is for solving the problem. Subsequently, the population is iteratively evolved for a certain number of generations thanks to the genetic operators (*crossover* and *mutation*) until the optimal solution is reached.

In this study, three tests were carried out [57], each of which had datasets containing race, sex and measurements of the *humerus*, *radius*, *ulna*, *femur*, *tibia* and *fibula*:

- The first test contained the six bone sizes of men and women of European-American origin;

- The second test contained the six bone sizes of African-American men and women;

- The third test contained all male and female data from the two previous tests (including race).

This choice was made to understand if the attributes such as sex and race really served to calculate the stature. Before being given "in the meal" to the algorithms, the data must pass a data processing phase in which they try to reduce the dimensionality and normalize (the final result must then be denormalized) the data by eliminating the characteristics that do not significantly influence the process of the estimate. In tests, the *principal component analysis* (PCA) [58] was used, reducing the size of the data to two. During the training phase the models are built and optimized:

- ANN is a feedforward neural network trained using the backpropagation-momentum learning principle [59], an algorithm useful for speeding up network convergence. It consists of three levels: the input level has $n_i$ neurons (equal to the dimensionality of the input vector) and a bias neuron, a single hidden level with $n_h$ neurons equal to the nearest integer given by $n_h = \sqrt{n_i \cdot n_o}$ (3) and finally a single neuron at the output level. In the beginning, the weights of the connections will be initialized randomly between [-0.5, 0.5] and only during training will they change to reach optimal values.

- For GA it is necessary to define a formula for the aforementioned height:

$$PS(s_i) = \sum_{k=1}^{m} w_k \cdot s_{ik} \quad (4)$$

and that of the fitness function

$$fitness(c) = \sum_{i=1}^{n} \left| \sum_{j=1}^{m} c_j \cdot s_{ij} - stature_i \right| \quad (5)$$

Twenty times *k-fold cross-validation* [58] is used to evaluate the performance and calculate the error in the learning phase of the ANN and GA models. After dividing the dataset into k folds, k-1 folds are used for training while the remainder is used for testing; everything is repeated for k times by varying the test fold. Furthermore, at each step of the process both the *average of the absolute errors* (MAE) [60]

$$MAE = \frac{1}{p} \sum_{i=1}^{p} |stature_i - out_i| \quad (6)$$

and the *standard error of the estimate* (SEE) [61]

$$SEE = \sqrt{\frac{1}{p} \sum_{i=1}^{p} (stature_i - out_i)^2} \quad (7)$$

are calculated to measure the accuracy of the model.

> *Key line on formula (4), (5), (6) e (7) notation:*
>
> - $m$: number of features;
> - $n$: number of training instance;
> - $s_{ik}$: measure k-th of the instance i after preprocessing;
> - $c_j$: j-th component of the vector.
> - $stature_i$: true stature of the instance i;
> - $out_i$: estimated stature of the instance i;
> - $p$: number of test samples.

The results obtained, considering all 20 cross-validations, are shown in the **Table 5**.

| Case of study | Model | Class | Measure | Min (cm) | Max (cm) | Mean (cm) | SD (cm) |
|---|---|---|---|---|---|---|---|
| First | ANN | Eu.-Am. males | MAE | 2,1746 | 3,0061 | 2,3918 | 0,1694 |
| | | | SEE | 2,9708 | 3,2765 | 3,1087 | 0,0967 |
| | | Eu.-Am. females | MAE | 2,8301 | 3,7833 | 3,1824 | 0,2186 |
| | | | SEE | 3,5054 | 4,3797 | 3,8859 | 0,243 |
| | GA | Eu.-Am. males | MAE | 2,1989 | 2,7786 | 2,5192 | 0,1828 |
| | | | SEE | 2,6111 | 3,3745 | 3,0565 | 0,2377 |
| | | Eu.-Am. females | MAE | 2,8011 | 4,3617 | 3,5917 | 0,3759 |
| | | | SEE | 3,084 | 4,8843 | 4,0973 | 0,4317 |
| Second | ANN | Af.-Am. males | MAE | 2,2617 | 2,9232 | 2,3307 | 0,1435 |
| | | | SEE | 2,8391 | 2,9758 | 2,9093 | 0,0401 |
| | | Af.-Am. females | MAE | 3,3741 | 4,2463 | 3,5447 | 0,178 |
| | | | SEE | 4,2463 | 4,4678 | 4,3797 | 0,0698 |
| | GA | Af.-Am. males | MAE | 2,4058 | 2,601 | 2,4832 | 0,0469 |
| | | | SEE | 3,0455 | 3,2182 | 3,1159 | 0,0439 |
| | | Af.-Am. females | MAE | 2,8803 | 3,9403 | 3,4436 | 0,2502 |
| | | | SEE | 3,3218 | 4,4203 | 3,9339 | 0,2846 |
| Third | ANN | Eu.-Am. males | MAE | 2,2349 | 3,1855 | 2,4097 | 0,198 |
| | | | SEE | 3,0201 | 3,2983 | 3,1744 | 0,0679 |
| | | Eu.-Am. females | MAE | 3,2036 | 4,1291 | 3,4562 | 0,2027 |
| | | | SEE | 3,9793 | 4,3685 | 4,1396 | 0,1199 |
| | | Af.-Am. males | MAE | 2,3066 | 3,0574 | 2,4523 | 0,1506 |
| | | | SEE | 2,8881 | 3,1721 | 3,0461 | 0,0758 |
| | | Af.-Am. females | MAE | 3,2441 | 4,1644 | 3,4172 | 0,1967 |
| | | | SEE | 3,98 | 4,4414 | 4,1225 | 0,1034 |
| | GA | Eu.-Am. males | MAE | 2,3736 | 2,8585 | 2,6734 | 0,1295 |
| | | | SEE | 2,7865 | 3,4546 | 3,225 | 0,1528 |
| | | Eu.-Am. females | MAE | 2,8977 | 4,0991 | 3,302 | 0,2727 |
| | | | SEE | 3,1736 | 4,7546 | 3,7397 | 0,3422 |
| | | Af.-Am. males | MAE | 2,5537 | 2,9871 | 2,7098 | 0,1042 |
| | | | SEE | 3,0997 | 3,6542 | 3,3079 | 0,1317 |
| | | Af.-Am. females | MAE | 2,8418 | 4,0111 | 3,4577 | 0,2762 |
| | | | SEE | 3,317 | 4,4031 | 3,9577 | 0,2508 |

**Table 5:** *Detailed study results that consider all 20 cross-validations.*

Both models performed well in stature estimation, the best results emerged in cases where only *bone measurements* were present (*sex* and *race* are not very relevant). In general, the ANN approach has obtained better average results than the GA: ANN, in addition to being more flexible, also allows to take into account the non-linearity between the data; GA, on the other hand, uses an easier to use linear formula. Even by performing the Wilcoxon signed-rank test, ANN and GA do not get substantial performance differences. The results obtained by ANN and GA could be improved by using more recent methods [62].

In the **Table 6**, it can be observed that the models presented have obtained very good performances also with other already existing approaches.

| Approach | Average MAE (cm) | Average SEE (cm) |
|---|---|---|
| ANN [57] | 2,6665 | 3,2985 |
| GA [57] | 2,8767 | 3,6697 |
| Anatomical approach to calculating height from clavicle in Punjabis. [63] | ------- | 32 |
| Anatomical approach to calculating height from the vertebral column in American blacks. [64] | ------- | 6,1 |
| Mathematical approach to calculating height from tibias. [65] | ------- | 4,81 |
| Mathematical approach to calculating height from long bones from Poland people. [66] | ------- | 4,85 |
| Mathematical approach to calculating height from somatometry of skull. [67] | ------- | 7,59 |
| Mathematical approach to calculating height from the long bones of the ancient Romans. [68] | ------- | > 6 |
| Anatomical approach to calculating height from hand and phalanges length. [69] | ------- | 4,58 |
| Mathematical approach to calculating height from measurements of the skulls of indigenous people from southern Africa. [70] | ------- | 5,31 |
| Anatomical approach to calculating height from ancient Egyptian skeletons. [71] | ------- | 3,05 |
| Mathematical approach to calculating height from radius length in Bengali men. [72] | ------- | 2,482 |
| Anatomical approach to calculating height with method of Fully. [73] | < 3,5 | ------- |
| Mathematical approach to calculating height from femur/stature ratio and "race". [74] | 2,78 | ------- |

**Table 6:** *Comparison table between this study and other important approaches in the literature.*

# 5. Algorithms of artificial intelligence for the identification of archeological sites

Over the millennia man has always changed the soil: to build houses or to harness the waters the ground has been continuously excavated to create trenches, pits, canals or simple holes which have affected the surface of the soil altering its shape. But since the most remote antiquity man has dug the ground also to look for the materials necessary for its subsistence: the clay to produce the vases, the minerals to extract the metal necessary for the tools and the stones to build buildings.

Subsequently, the interest of seeking in the ground also extended to what man himself had placed there: the earth has been enriched with "treasures" that have become part of our collective imagination. The very concept of "treasure" has gradually changed over time, also based on the different value that the various civilizations have attributed from time to time to certain objects. For example, in Eastern civilizations the rediscovery of ancient palaces and dynastic sanctuaries served to consolidate and increase the importance of the genealogical tree from which the sovereign descended, in Greece the rediscovery of the tombs and weapons of heroes was a symbol of worship and re-enactment of the past. During the Middle Ages, churches were embellished with ancient artefacts recovered and purified with prayers specially dedicated "to the vases discovered in ancient places". The land, however, was not yet seen as a potential custodian of a historical tale.

If the search for precious objects has therefore been a habit long cultivated by man, archeological excavation can be considered one of the most recent forms of this trend. The rise of the archeological excavation, as we understand it today, occurs in the nineteenth century with archaeology intended as historical science. The first archeologist to exhibit his practical method of excavation was Boni in a famous essay of 1901 [75]. The method was applied for the first time in the excavations of the Roman Forum and consisted in the execution of stratigraphic surveys aimed at recognizing the stratifications and in the removal of the individual layers "according to their natural deposit". The historical interpretations that Boni said of his excavations show that, at the turn of the twentieth century, the fundamental concepts of the stratigraphic investigation of the terrain were now acquired and could also be applied in the practice of monumental excavation.

In addition to all the traditional methods of detecting archeological and excavation sites, some based on artificial intelligence have recently been developed relating to archeological sites not only terrestrial, but also marine. The following will analyze the artificial intelligence techniques used in the search for archaeological sites in Brandenburg [76], the "princely" tombs in the Eurasian steppe [77], the urban area of the ancient city of Falerii Novi [78] , the monumental site of Kuelap in Peru [79] and the exploration of marine archaeological sites [80], [81].

## 5.1 Identification of terrestrial archeological sites

*Archaeological predictive modelling* (APM) deals with developing methods suitable for the identification of potential archaeological sites in geographic space, using regression and classification techniques, spatial statistics and heuristic approaches. *Geographic information* *system* (GIS) is a fundamental part of the APM because it provides tools for the storage and analysis of spatial variables as well as the interface between geo-archaeological information and advanced numerical processing at the base of all archaeological predictive models. For this type of operation, it is assumed that the settlements have environmental characteristics: characteristics of the conformation of the soil (height, slope and aspect angle), the distance from surface waters, the level of groundwater, the structure and the quality of the soil.

The artificial intelligence technique used for this purpose is the *Artificial Neural Networks* (ANNs) [55]. The artificial intelligence technique used for this purpose is the Artificial Neural Networks (ANNs). ANNs are a computational model consisting of artificial "neurons" and are inspired by a biological neural network; they are often used for classification problems. In the context of APM, ANNs are widely used, thanks to their robustness and their training and testing method, because they effectively manage the most serious problems that usually afflict archaeological data: missing or incomplete data, noisy data (for example in this case data that do not refer mainly to environmental factors, but to political, religious or social reasons that led to the settlement in that place) and finally to the non-linear relationships between the data. The most used ANNs variants are:

- *Multilayer Perceptron Network* (MLP) [82] is one of the basic architectures of the ANNs. It consists of multiple layers of nodes (an input layer, 1 or more hidden layers and an output layer), in which each layer is completely connected to the next. This network provides a supervised learning technique called backpropagation: during the training, the values of the connections are updated until the optimal value is found. In the case of APM, the input consists of a vector containing the environmental characteristics described above, while the output will consist of two nodes: "site present" and "no site present".

- *Probabilistic Neural Network* (PNN) [83] is a special architecture of ANNs with a kernel density estimation, a statistical method that uses functions to approximate the behavior of a sampling distribution [84]; for this reason the network can estimate the true probability function according to the input data. In a PNN, the operations are organized in a multilayer feedforward network with four levels: an input layer, a model layer, a summation layer and finally an output layer. In this case the only probability that there is a site can be output.

- *Convolutional Neural Network* (CNN) [32] (**Figure 23**) is a family of neural networks widely used in the field of computer vision and, more generally, with data that have spatial relationships. It is a feedforward ANN in which the way neurons are connected is very reminiscent of the organization of the animal visual cortex. The structure of a CNN is organized in layers of three different types: the convolutional layer, the pooling layer and finally the fully connected one. They accept input data similar to a grid or that can be treated as such, for example images.

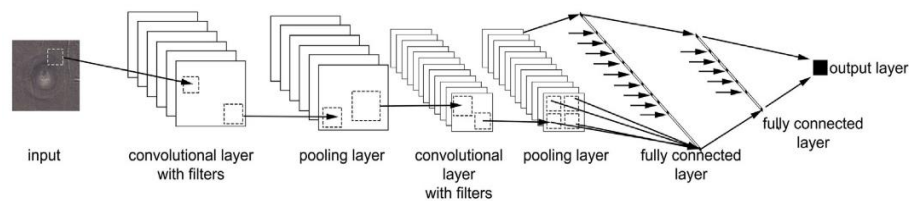

**Figure 23:** *Example of architecture of a Convolutional Neural Network (CNN).*

- *Self-Organizing Feature Map* (SOM) [85] is a type of artificial neural network that is trained through unsupervised learning to produce a low-dimensional (typically two-dimensional) discrete representation of the input space of training samples, called a map. SOMs differ from other artificial neural networks in that they apply competitive learning with respect to error-correction learning and use a neighbourhood function to preserve the topological properties of the input space. SOM is useful for visualization because it creates low-dimensional views of high-dimensional data, like multidimensional scaling.

In the *Archäoprognose Brandenburg project* [76], an attempt was made to create an APM tool that identifies the location of possible archaeological sites in Brandenburg in north-east Germany. *Ainet*, a free software that combines features of the PNNs and SOMs architectures, was used, offering the user a graphical interface for the input data.

Instead, in the *Dzungaria Landscape project* [77], a CNN was used to find "princely" tombs in the Eurasian steppe (an area difficult to examine due to its vastness and therefore under different administrations, languages and institutions). At the foot of the Chinese Altai mountains, in particular in the Heiliutan area, an intense activity of building funeral mounds dating back to the Iron Age has been identified. To educate the *CNN network*, it has been decided to focus on the "Saka" architecture because it presents homogeneity: 59 large elevated mounds built with a mixture of pebbles and surrounded by circular rings of large stones or circular ditches were found by satellites. Both methods have had good site estimation results.

Two other noteworthy projects are:

- The first [78] uses the GPR (Ground Penetrating Radar) to rebuild an entire buried urban area of the ancient city of Falerii Novi (located about 50 km north of Rome and founded in 241 BC). The choice of this city is due to the fact that, in addition to already having information from the excavations of the nineteenth and twentieth centuries together with an epigraphic and sculptural analysis, it was among the first Roman cities subjected to a complete survey with the gradiometer at vertical flow which has given a very clear plan of the whole site and which has made it possible to frame the urban plan within those typical of contemporary Roman colonies. The information will be used as feedback and an integration of the new GPR acquisitions. In recent decades, archaeological literature has built a reconstructive picture of urban planning based on the large excavation campaigns that have affected centers such as Ostia and Pompeii which, however, certainly do not represent the great variety of urban experiences in the Roman world. Therefore, studies of this type can bring to light new information on the variety of urban experiences of the Roman world, without necessarily carrying out excavations. The GPR and magnetometric investigation techniques are among the most popular methodologies for prospecting the first meters of subsoil in search of underground structures: the first exploits the reflections of the electromagnetic waves transmitted in the ground or in the structure to be examined during the transitions between materials of different dielectric permittivity and returns high resolution 3D images of the structures identified. Instead, the second measures the total intensity of the Earth's magnetic field and the variations caused by structures with magnetic properties different from those of the medium that incorporates them. In Italy, since 2015, large-scale GPR analysis has started, but the amount of data becomes very large and this takes a long time to perform a complete visual analysis and manual digitization of anomalies; for this reason, the researchers proposed the merging of different geophysical data sets (in this case, the visualization of the GPR detection data and magnetometry were merged into a single image, assigning the two data sets of different color channels, called also RGB composition). Finally, the team of researchers also thought of applying this procedure to other cities, taking into account that there may be less favorable sites than this: in fact, often the techniques of analysis of the territory depend on the type of soil from the geology and archaeological conservation of each place. The data from this study were uploaded to an open access online archive with the Archeology Data Service so that it can be viewed and used by anyone interested in it.

- The second [79] is the studies carried out at the monumental site of Kuelap in Peru thanks to LiDAR and artificial intelligence. The site was inhabited by the Chachapoyas (also called *cloud warriors*) until the Incas conquered their land a few years before the arrival of the Spaniards around the XV century. It is located in the colonial city of Chachapoyas, in the Amazon region; after having undergone centuries of abandonment, the rediscovery of the place was thanks to the news of extraordinary archaeological remains that spread with the investigation of a dispute over land discovered by a magistrate of the city, who described the monumental settlement now known as *"La Fortaleza"* in 1843. On the site, in addition to the immense "*fortress*" (with external walls that reach 20 meters in height) called Kuelap, there are also a surprising number of stone settlements in ruins (over 700 circular houses that they extend over 5km2), many of which have monumental proportions and houses richly decorated with lithic sculptures. Although studied sporadically since 1893, the first real and important results that form the basis of the chronology of the site are those made by Ruiz Estrada in the late sixties. Four smaller sites were also discovered (*Muralla de Malcapampa, La Barreta, Pampa Linda, Imperio and Quiúquita-Lahuancho*) with various aspects that differentiate them, but which seem to be connected to La Fortaleza based on their proximity and distribution on the slopes at Northeast and South of the main monument. Today they are covered in dense mountain forest vegetation, as all efforts to prepare Kuelap for tourism have focused on the main site. The area under study is particularly impervious, with a particularly humid and rainy climate and widely covered by dense vegetation which, in fact, prevents its understanding and interpretation using exclusively photogrammetric information. In the present experience, original methodologies have been implemented that allow to obtain, starting from the set of available data sets, a highly automated mapping, using both suitably customized standard processing techniques, and deep learning techniques for the identification and classification of the traces identified. During the LiDAR and photogrammetric survey, the production of DSM (*Digital Surface Model*) and DTM (*Digital Terrain Model*) was performed with a resolution of 15 cm and orthomosaic with a resolution of 5 cm. Subsequently, the filtering and classification phases of the point cloud were carried out using macros specially developed in the ATLAS environment (www.theatlasgis.com, MEDS property) made due to the particular morphology of the objects to be detected (Noise caused by high humidity, dense occlusive vegetation, very small size of the buildings present, ...). The primary objective of this study was to verify the ability of this technology to produce images of the site surface which could adequately penetrate the vegetation and which produced a sufficiently accurate representation of the morphology of the soil in order to safely identify the buildings. First of all, the relief of the city of Kuelap has been manually mapped and 84 buildings have been identified that the software must learn to know, selecting them in order to cover diversified situations. After identifying the shape, centroid and approximate size for each building, sections of LiDAR relief centered on the building were extracted and variants were created, in order to disturb the data set by inserting translations and rotations, and altering the set of points with random noise. This resulted in a total of 125 variants for each building, bringing the total set of samples to 10500. 70% of the samples were used to train the TensorFlow-based deep learning software [39] and implemented in C ++. for the test the remaining 30%. This technology offered an unprecedented, non-destructive means of identifying and provisionally characterizing unknown and undervalued peculiarities of the surrounding built landscape, pending subsequent checks. This can never replace the traditional observations on the ground of archaeologists, the two operations are complementary because both are necessary to carefully analyze the archaeological sites.

## 5.2 Venus, Virtual ExploratioN of Underwater Sites

VENUS [80], [81] is a project funded by the European Community, aimed at the virtual exploration of marine archeological sites, such as shipwrecks. It started on July 1st 2006, lasted 36 months and, at its end, all the details and results of the project were collected at the following web address: https://cordis.europa.eu/project/id/034924.

The VENUS project has the main purpose of providing scientific methodologies and technological tools for the virtual exploration of marine archeological sites. Virtual exploration allows both experts and the public to see and study sites, very often, located at high depths: darkness, low temperature and oxygen percentage are important factors for the conservation of wrecks, they also cause the impossibility of a direct study by archeologists. Marine sites are not always accessible by divers, but most of the time the use of AUVs / ROVs (Autonomous Underwater / Remotely Operated Vehicles) is required. These devices allow an automatic collection of high-quality data, as well as an improvement in terms of efficiency, economy and safety. In addition, these sites are damaged by trawling which jeopardizes not only their readability but also destroys the entire surface layer; for this reason, the VENUS project is very important. The project aims to drastically change this situation, making marine sites practically accessible to everyone and offering tools to archeologists to collect and organize the archeological data coming from the shipwreck.

This project also includes three in-depth investigations of different marine archeological sites of shipwrecks in excellent condition and in different sea conditions to apply the entire VENUS platform and show some possible immersive VR applications:

- The first is a shipwreck site (**Figure 24**), near the island of *Pianosa* (in Tuscany, Italy) at about 36m depth. Around a hundred Roman amphorae have been discovered in an area of 35m × 38m, it is probably the site of more than one shipwreck. The island was the site of the exile of Marco Vipsanio Agrippa, nephew of the emperor Augustus, as evidenced by the remains of a Roman villa found on the island.

- The second consists of two sites in Portugal, one on the south coast (*Algarve*) and one on the east coast near *Troia* (a Roman industrial complex where fish and wine were processed and used locally produced amphorae). Both are located at a depth of about 50m and have numerous amphorae.

- The last site (**Figure 25**), called "*Calanque de Port Miou*", is among the most beautiful and best-preserved shipwrecks in the Marseille area; the site has a large mound of Roman wine amphorae still intact and visible from a depth of 22m even if it is located at a depth of 120m.

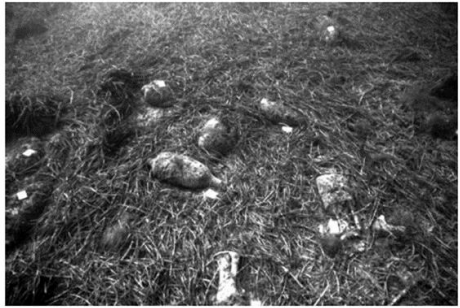

**Figure 24:** A *view from the site near Pianosa.*

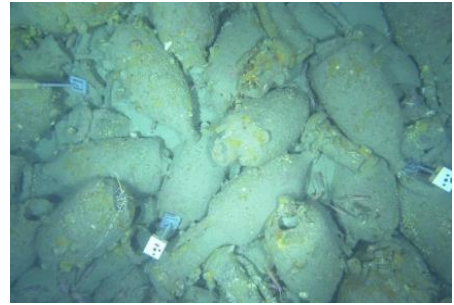

**Figure 25:** A *photo taken by an ROV of the "Calanque de Port Miou" site.*

During the three investigations, efforts were made to formalize a method that respected the following objectives:

1. *Underwater exploration best practices and procedures*

In this objective, the best procedures are defined to collect data efficiently, economically and safely from archeological sites:

- Efficiency because a good amount of data is collected automatically to create a satisfactory virtual reconstruction;

- Economy because "ready-to-use" equipment is used, which archeologists can use easily and with few adaptations;

- Safety because the risks of men working in hostile environments are reduced.

The technology identified as necessary to satisfy the previous points is the use of Uninhabited Underwater Vehicles (UUVs), Autonomous Underwater and Remotely Operated Vehicles (AUVs / ROVs) [86]. UUVs are the "instrument par excellence" for this task because they are equipped with optical, acoustic and magnetic sensors and can recover good quality data from the site (**Figure 26**).

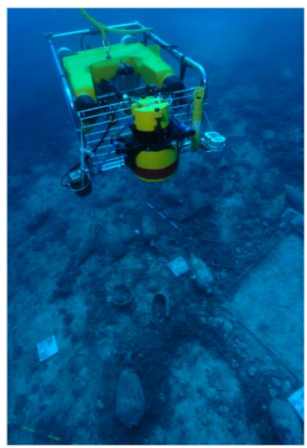

**Figure 26:** *ROV which collects marine site data.*

2. *Underwater 3D survey merging optic and acoustic sensors*

This second objective seeks to define practical procedures for merging optical and acoustic data into a coherent representation. To make data collection more accurate and facilitate the construction of the virtual environment, in addition to the sensors already described above, the vehicles will be equipped with navigation systems so as to geo-reference the data. Simple photogrammetric techniques will be created to allow the use of software originally conceived for terrestrial, obviously after appropriate image processing. A semi-automatic surface densification technique will be implemented to generate a large quantity of 3D points, based on the correlation of the multiple images collected and to simplify the photogrammetric process during the printing phase. Finally, a digitization tool will be developed that will offer archeologists the possibility of making 3D measurements starting from the product of the merger operation: a 3D mesh and a series of photographs oriented in the same reference system.

3. *Managing and revising archeological knowledge*

The aim of the third objective is to provide software tools to archeologists in order to efficiently process and manage the data obtained from the previous point. The data collected very often require the integration of scholars' knowledge and, moreover, they may be incomplete, uncertain and may conflict with others. For this reason, artificial intelligence methods and other tools that allow to adequately represent archeological information will be researched, to review and merge it respecting the following principles:

- To design and build specific ontologies for each type of finding identified, examining it and comparing it with others;

- To design formalisms suitable for representing knowledge on the basis of the information and data collected;

- To select which merger strategy is best suited to the nature of the study being conducted on the basis of the reliability and quality of the data;

- To evaluate the flexibility of the merger;

- To define reversible operations of fusion: existing fusion operations in artificial intelligence are not reversible, but in real cases (such as this one) it is necessary to define it.

4. *Mixed reality modelling*

In the fourth objective it was attempted to create the archeological site in a virtual universe starting from the digital model produced in the previous point and to allow archeologists to work on it in the most natural way possible thanks to virtual and augmented reality. In addition to showing the site, the software could also offer tools that allow the reconstructions of artefacts found on the site according to the hypotheses of scholars. The possibility of simultaneous exploration by multiple researchers and access to the various hardware levels depending on the skills of each researcher are also very useful. From the last idea follows the need to create many demonstrators using the same engine:

- An immersive demonstrator that uses only immersive devices, such as the VR helmet and gloves;

- A semi-immersive demonstrator that uses stereo displays and gloves or a 3D joystick for interaction;

- A low-level platform that uses the standard devices of a desktop computer.

# 6. Conclusion and further works

All the methods exposed aim to simplify and speed up all those laborious and complex operations that archaeologists must perform; their application has obtained very satisfactory results that encourage us to continue studying and research, not only for the improvement of existing methods but also for the formulation of new ones:

- The Minerva software (**section 2.2**) will improve the communication and negotiation aspects of the interaction between agents and the automatic retrieval of information regarding works of art from the web. In addition, programmers will aim to validate the adaptability of its architecture in different contexts: the same Minerva kernel will be applied to the organization of other museums and collections, where different criteria are used during the organization process.

- As regards the ArchAIDE project (**section 3.1.1**), in addition to having implemented and tested the automatic recognition system on mobile devices and on desktop platforms, the software will be improved so that it is easy to use also in the field and that the data collected by different devices are as homogeneous as possible. Furthermore, programmers will try to integrate the data collected by users during the recognition process in the field with those coming from the digitization process in order to create a database that is as uniform as possible.

- In the reconstruction of the ostraka (**section 3.1.2**), programmers will try to decrease the incorrect alignments, providing the system with metadata information, for example: the material, the language, the type of ceramic etc. In addition, an interactive interface will be provided that allows archaeologists to manually select the best reconstruction proposal.

- PHYTHIA and the processing pipeline of PHI-ML (**section 3.2**) have been published as open source software so that it can help future research and inspiration for upcoming interdisciplinary works.

- Future changes on the study of sex starting from the skull (**section 4.1**) could allow the use of data from the archaeological environment and can also be applied to other parts of the skeleton with dysmorphic characters such as, for example, the pelvis. Furthermore, the source of the inputs to the network could be changed: not only the CT scans but also images taken by the camera or smartphone in order to offer rapid use in the field.

- Miletus in Turkey and Cyrene in Libya will be the backdrop for other studies carried out with GPR and magnetometry as for Falerii Novi (**section 5.1**).

- The study carried out on the monumental site of Kuelap (**section 5.1**) will be extended to the neighboring sites in order to better understand the relationship between them and Fortaleza; for most people visiting the archaeological site is difficult (many hours by car and walking on steep terrain), as well as for the current planetary health emergency, some researchers are working to create a 3D reconstruction of the site in order to encourage online virtual tourism.

- The researchers of the VENUS project (**section 5.2**) will undertake to provide increasingly precise 3D virtual maps of marine archaeological sites and user-friendly interfaces both for the study by archaeologists and for the exploration of anyone who wishes to visit the site. Open source software and results will also be published to encourage the dissemination and improvement of the project.

The technological development, exposed in Moore's law, puts an increasingly powerful hardware disposition in the processing of large quantities of data and their conservation; in addition, the use of other devices (for example smartphones, holographic techniques, cloud servers), in addition to the traditional computer, has given rise to new approaches to study and interaction with archaeology.

## 7. References

[1] "Archaeology - Wikipedia." [Online]. Available: https://en.wikipedia.org/wiki/Archaeology. [Accessed: 16-Apr-2020].

[2] "INFORMATICA ARCHEOLOGICA in 'Enciclopedia Italiana.'" [Online]. Available: http://www.treccani.it/enciclopedia/informatica-archeologica_%28Enciclopedia-Italiana%29/. [Accessed: 16-Apr-2020].

[3] "CAA International." [Online]. Available: https://caa-international.org/. [Accessed: 16-Apr-2020].

[4] "Archeologia e Calcolatori." [Online]. Available: http://www.archcalc.cnr.it/. [Accessed: 16-Apr-2020].

[5] M. B. Rosson and Association for Computing Machinery., Proceedings of the SIGCHI Conference on Human Factors in Computing Systems. ACM, 2007.

[6] W. Burgard et al., "Experiences with an interactive museum tour-guide robot," 1999.

[7] K. Fukui and O. Yamaguchi, "Facial feature point extraction method based on combination of shape extraction and pattern matching," Syst. Comput. Japan, vol. 29, no. 6, pp. 49–58, Jun. 1998, doi: 10.1002/(SICI)1520-684X(19980615)29:6<49::AID-SCJ5>3.0.CO;2-L.

[8] Y. Kuno et al., "The 15th IEEE International Symposium on Robot and Human Interactive Communication (RO-MAN06) Museum Guide Robot with Communicative Head Motion."

[9] E. Frontoni, Vision based mobile robotics : mobile robot localization using vision sensors. Lulu Com, 2012.

[10] D. Fox, W. Burgard, S. Thrun, and A. B. Cremers, "Position estimation for mobile robots in dynamic environments," Proc. Natl. Conf. Artif. Intell., pp. 983–988, 1998.

[11] H. P. Moravec, "Sensor Fusion in Certainty Grids for Mobile Robots," Sens. Devices Syst. Robot., vol. 9, no. 2, pp. 253–276, 1989, doi: 10.1007/978-3-642-74567-6_19.

[12] S. Thrun, W. Burgard, and D. Fox, "A Probabilistic Approach to Concurrent Mapping and Localization for Mobile Robots," Auton. Robots, vol. 5, no. 3–4, pp. 253–271, 1998, doi: 10.1023/A:1008806205438.

[13] D. Fox, W. Burgard, and S. Thrun, "The Dynamic Window Approach to Collision Avoidance," Robot. Autom. Mag. IEEE, vol. 4, pp. 23–33, Apr. 1997, doi: 10.1109/100.580977.

[14] S. Thrun, B. Arno, and T. Fr, "Map Learning and High-Speed Navigation in RHINO."

[15] D. Thomas and B. Mark, "An Analysis of Time-Dependent Planning," Proceeding Seventh Natl. Conf. Artif. Intell., vol. AAAI-88, pp. 49–54, 1988.

[16] B. Richard, Dynamic Programming, Sixth prin. Princeton, New Jersey: Princeton University Press, 1957.

[17] R. A. Howard, Dynamic programming and Markov processes. Oxford, England: John Wiley, 1960.

[18] H. J. Levesque, R. Reiter, Y. Lespérance, F. Lin, and R. B. Scherl, "Golog: A logic programming language for dynamic domains," J. Log. Program., vol. 31, no. 1–3, pp. 59–83, Apr. 1997, doi: 10.1016/S0743-1066(96)00121-5.

[19] F. Amigoni, V. Schiaffonati, and M. Somalvico, "Minerva: An Artificial Intelligent System for Composition of Museums."

[20] F. Amigoni and V. Schiaffonati, "A New Version of Minerva for Organizing Archeological Museums."

[21] F. A. Politecnico, D. Milano, F. Amigoni, and V. Schiaffonati, "The Minerva Multiagent System for Supporting Creativity The Minerva Multiagent System for Supporting Creativity in Museums Organization," 2003.

[22] E. Friedman-Hill, "Jess, The Java Expert System Shell," Biosystems, Jan. 2003.

[23] G. Caggianese, G. De Pietro, M. Esposito, L. Gallo, A. Minutolo, and P. Neroni, "Discovering Leonardo with artificial intelligence and holograms: A user study," Pattern Recognit. Lett., 2020.

[24] J. C. Sprott, Physics demonstrations : a sourcebook for teachers of physics. University of Wisconsin Press, 2006.

[25] W. Y. (William Y. Adams and E. W. (Ernest W. Adams, Archaeological typology and practical reality : a dialectical approach to artifact classification and sorting. Cambridge University Press, 2008.

[26] J. A. Barceló, "Computational intelligence in archaeology," Comput. Intell. Archaeol., vol. 25, no. 2003, pp. 1–418, 2008, doi: 10.4018/978-1-59904-489-7.

[27] A. Salazar, G. Safont, L. Vergara, and E. Vidal, "Pattern recognition techniques for provenance classification of archaeological ceramics using ultrasounds," Pattern Recognit. Lett., vol. 135, pp. 441–450, 2020, doi: 10.1016/j.patrec.2020.04.013.

[28] C. Romanengo, S. Biasotti, and B. Falcidieno, "Recognising decorations in archaeological finds through the analysis of characteristic curves on 3D models," Pattern Recognit. Lett., vol. 131, pp. 405–412, 2020, doi: 10.1016/j.patrec.2020.01.025.

[29] M. L. Gualandi et al., "ArchAIDE-Archaeological Automatic Interpretation and Documentation of cEramics."

[30] G. Gattiglia, "CLASSIFICARE LE CERAMICHE: DAI METODI TRADIZIONALI ALL'INTELLIGENZA ARTIFICIALE. L'ESPERIENZA DEL PROGETTO EUROPEO ARCHAIDE," Archeol. QUO VADIS? Riflessioni Metodol. sul Futur. di una Discip. Atti, p. 442, 2018.

[31] F. Anichini and G. Gattiglia, "Big Archaeological Data. The ArchAIDE project approach," in GARR-Conf17-proceedings-03 Big, 2018, p. 3.

[32] Y. Lecun, L. Bottou, Y. Bengio, and P. Ha, "Gradient-Based Learning Applied to Document Recognition," Proc. IEEE, no. November, pp. 1–46, 1998, doi: 10.1109/5.726791.

[33] V. Mayer-Schönberger and K. Cukier, Big data : a revolution that will transform how we live, work, and think. Houghton Mifflin Harcourt, 2013.

[34] R. Schutt and C. O'Neil, "Introduction: What Is Data Science?," Doing Data Sci. Straight Talk From Front., pp. 1–15, 2013.

[35] R. Smith, "An Overview of the Tesseract OCR Engine," Ninth Int. Conf. Doc. Anal. Recognit. (ICDAR 2007), vol. vol.2, pp. 629–633, 2007, doi: 10.1007/978-3-642-22027-2_57.

[36] M. Potenziani, M. Callieri, M. Dellepiane, M. Corsini, and R. Scopigno, "3DHOP: 3D Heritage Online Presenter," Comput. Graph., vol. 52, 2015, doi: 10.1016/j.cag.2015.07.001.

[37] C. Ostertag and M. Beurton-Aimar, "Matching ostraca fragments using a siamese neural network," Pattern Recognit. Lett., vol. 131, pp. 336–340, Mar. 2020, doi: 10.1016/j.patrec.2020.01.012.

[38] P. De Smet, "Reconstruction of ripped-up documents using fragment stack analysis procedures," Forensic Sci. Int., vol. 176, no. 2–3, pp. 124–136, Apr. 2008, doi: 10.1016/j.forsciint.2007.07.013.

[39] "TensorFlow." [Online]. Available: https://www.tensorflow.org/. [Accessed: 13-Mar-2020].

[40] "Home - Keras Documentation." [Online]. Available: https://keras.io/. [Accessed: 13-Mar-2020].

[41] Y. Assael, T. Sommerschield, and J. Prag, "Restoring ancient text using deep learning: a case study on Greek epigraphy," Oct. 2019.

[42] "PHI Greek Inscriptions." [Online]. Available: https://inscriptions.packhum.org/. [Accessed: 19-Mar-2020].

[43] "Open Greek and Latin Project | Digital Humanities." [Online]. Available: https://www.dh.uni-leipzig.de/wo/projects/open-greek-and-latin-project/. [Accessed: 19-Mar-2020].

[44] D. A. Smith, J. A. Rydberg-Cox, and G. Crane, "The Perseus Project: a digital library for the humanities," Lit. Linguist. Comput., vol. 15, 2000, doi: 10.1093/llc/15.1.15.

[45] I. Sutskever, O. Vinyals, and Q. V. Le, "Sequence to sequence learning with neural networks," Adv. Neural Inf. Process. Syst., vol. 4, no. January, pp. 3104–3112, 2014.

[46] S. Hochreiter and J. Schmidhuber, "Long Short-Term Memory," Neural Comput., vol. 9, no. 8, pp. 1735–1780, 1997, doi: 10.1162/neco.1997.9.8.1735.

[47] M.-T. Luong, H. Pham, and C. D. Manning, "Effective Approaches to Attention-based Neural Machine Translation," Conf. Proc. - EMNLP 2015 Conf. Empir. Methods Nat. Lang. Process., pp. 1412–1421, Aug. 2015.

[48] "Studi Anatomici di Leonardo Da Vinci." [Online]. Available: https://www.leonardodavinci-italy.it/anatomia/anatomia-umana. [Accessed: 12-Mar-2020].

[49] J. Bewes, A. Low, A. Morphett, F. D. Pate, and M. Henneberg, "Artificial intelligence for sex determination of skeletal remains: Application of a deep learning artificial neural network to human skulls," J. Forensic Leg. Med., vol. 62, pp. 40–43, Feb. 2019, doi: 10.1016/j.jflm.2019.01.004.

[50] C. Szegedy et al., "Going deeper with convolutions," Proc. IEEE Comput. Soc. Conf. Comput. Vis. Pattern Recognit., vol. 07-12-June, pp. 1–9, 2015, doi: 10.1109/CVPR.2015.7298594.

[51] "Pretrained GoogLeNet convolutional neural network - MATLAB googlenet - MathWorks Italia." [Online]. Available: https://it.mathworks.com/help/deeplearning/ref/googlenet.html;jsessionid=6e2403278b76bcd1b70f7c832bb6. [Accessed: 24-Feb-2020].

[52] D. T., "Methods of estimating height from parts of skeleton," Med. Reconstr. New York, vol. 46, pp. 293–296, 1884.

[53] K. Pearson, "IV. Mathematical contributions to the theory of evolution.—V. On the reconstruction of the stature of prehistoric races," Philos. Trans. R. Soc. Ser., vol. 192, pp. 169–244, 1899.

[54] G. Czibula, V. S. Ionescu, D. L. Miholca, and I. G. Mircea, "Machine learning-based approaches for predicting stature from archaeological skeletal remains using long bone lengths," J. Archaeol. Sci., vol. 69, pp. 85–99, May 2016, doi: 10.1016/j.jas.2016.04.004.

[55] J. Feldman and R. Rojas, Neural Networks: A Systematic Introduction. Springer Berlin Heidelberg, 2013.

[56] M. Mitchell, An Introduction to Genetic Algorithms. Bradford Books, 1998.

[57] R. J. Terry, "Terry Collection Postcranial Osteometric Database," 2015. [Online]. Available: http://anthropology.si.edu/cm/terry.htm.

[58] D. G. S. Richard O. Duda, Peter E. Hart, Pattern Classification, 2nd Edition. John Wiley and Sons, 2001.

[59] P. N. S. Russell, "Artificial Intelligence - a Modern Approach, Prentice Hall International Series in Artificial Intelligence," Prentice Hall, 2003.

[60] "Mean absolute error - Wikipedia." [Online]. Available: https://en.wikipedia.org/wiki/Mean_absolute_error. [Accessed: 17-Feb-2020].

[61] "Standard error - Wikipedia." [Online]. Available: https://en.wikipedia.org/wiki/Standard_error#Estimate. [Accessed: 17-Feb-2020].

[62] J. Del Ser et al., "Bio-inspired computation: Where we stand and what's next," Swarm Evol. Comput., vol. 48, no. April, pp. 220–250, 2019, doi: 10.1016/j.swevo.2019.04.008.

[63] B. SINGH and H. S. SOHAL, "Estimation of stature from clavicle in Punjabis; a preliminary report.," Indian J. Med. Res., vol. 40, no. 1, pp. 67–71, Jan. 1952.

[64] G. L. Tibbetts, "Estimation of Stature from the Vertebral Column in American Blacks," J. Forensic Sci., vol. 26, no. 4, p. 11427J, Oct. 1981, doi: 10.1520/jfs11427j.

[65] T. D. Holland, "Estimation of Adult Stature from Fragmentary Tibias," J. Forensic Sci., vol. 37, no. 5, p. 13309J, Sep. 1992, doi: 10.1520/jfs13309j.

[66] J. Kozak, "Stature Reconstruction From Long Bones . the Estimation of the Usefulness of Some Selected Methods for Skeletal Populations," Var. Evol., vol. 5, pp. 83–94, 1996.

[67] M. Chiba and K. Terazawa, "Estimation of stature from somatometry of skull," Forensic Sci. Int., vol. 97, no. 2–3, pp. 87–92, Nov. 1998, doi: 10.1016/S0379-0738(98)00145-5.

[68] G. K. Goldewijk and J. Jacobs, "The relation between stature and long bone length in the Roman Empire."

[69] O. P. Jasuja and Singh, "Estimation of stature from hand and phalange length," J Indian Acad Forensic Med, vol. 26, 2004.

[70] I. Ryan and M. Bidmos, "Skeletal height reconstruction from measurements of the skull in Indigenous South Africans," Forensic Sci. Int., vol. 167, pp. 16–21, 2007, doi: 10.1016/j.forsciint.2006.06.003.

[71] M. H. Raxter, C. B. Ruff, A. Azab, M. Erfan, M. Soliman, and A. El-Sawaf, "Stature estimation in ancient Egyptians: A new technique based on anatomical reconstruction of stature," Am. J. Phys. Anthropol., vol. 136, no. 2, pp. 147–155, Jun. 2008, doi: 10.1002/ajpa.20790.

[72] D. Champa Pal and A. K. Datta, "Estimation of stature from radius length in living adult Bengali males," Indian J. Basic Appl. Med. Res., vol. 3, no. 2, pp. 380–389, 2014.

[73] M. Raxter, B. Auerbach, and C. Ruff, "Revision of the Fully Technique for Estimating Statures," Am. J. Phys. Anthropol., vol. 130, pp. 374–384, Jul. 2006, doi: 10.1002/ajpa.20361.

[74] M. R. Feldesman and R. L. Fountain, "'Race' specificity and the femur/stature ratio," Am. J. Phys. Anthropol., vol. 100, no. 2, pp. 207–224, Jun. 1996, doi: 10.1002/(SICI)1096-8644(199606)100:2<207::AID-AJPA4>3.0.CO;2-U.

[75] G. Boni, Il metodo negli scavi archeologici, vol. IV, serie, no. XCIV. Arbor Sapientiae, 1901.

[76] B. Ducke and A. Brandenburg, "Archaeological Predictive Modelling in Intelligent Network Structures," 2003.

[77] G. Caspari and P. Crespo, "Convolutional neural networks for archaeological site detection – Finding 'princely' tombs," J. Archaeol. Sci., vol. 110, Oct. 2019, doi: 10.1016/j.jas.2019.104998.

[78] L. Verdonck, A. Launaro, F. Vermeulen, and M. Millett, "Ground-penetrating radar survey at Falerii Novi: a new approach to the study of Roman cities," Antiquity, vol. 94, no. 375, pp. 705–723, 2020, doi: 10.15184/aqy.2020.82.

[79] G. Righetti, S. Serafini, F. B. Rueda, W. Church, and G. Garnero, "Sotto Nuvole , sotto la Foresta : applicazioni Tecnologiche Lidar e di Intelligenza Artificiale per Nuove prospettive nel Sito monumentale di Kuelap - Perù," Archeomatica, pp. 6–13, 2020.

[80] P. Chapman et al., "VENUS, Virtual ExploratioN of Underwater Sites," 2006.

[81] R. Jeansoulin and O. Papini, "UNDERWATER ARCHAEOLOGICAL KNOWLEDGE ANALYSIS AND REPRESENTATION IN THE VENUS PROJECT : A PRELIMINARY DRAFT."

[82] E. B. Baum, "On the capabilities of multilayer perceptrons," J. Complex., vol. 4, no. 3, pp. 193–215, 1988, doi: 10.1016/0885-064X(88)90020-9.

[83] D. J. C. Mackay', "A Practical Bayesian Framework for Backpropagation Networks."

[84] M. D. Richard and R. P. Lippmann, " Neural Network Classifiers Estimate Bayesian a posteriori Probabilities ," Neural Comput., vol. 3, no. 4, pp. 461–483, Dec. 1991, doi: 10.1162/neco.1991.3.4.461.

[85] T. Kohonen, "Self-organized formation of topologically correct feature maps," Biol. Cybern., vol. 43, no. 1, pp. 59–69, Jan. 1982, doi: 10.1007/BF00337288.

[86] A. Pascoal, C. Silvestre, and P. Oliveira, "Vehicle and Mission Control of Single and Multiple Autonomous Marine Robots," Adv. Unmanned Mar. Veh., 2005, doi: 10.1049/PBCE069E_ch17.